%% file: main.tex
\documentclass{article} %
\usepackage{includes/iclr2024_conference,times}
\usepackage{enumitem}
\usepackage{graphicx}
\usepackage{booktabs} %
\usepackage{todonotes}
\usepackage{multirow}
\usepackage{csquotes}
\usepackage{caption}
\usepackage{array}
\usepackage{subcaption}
\usepackage{adjustbox}

\usepackage{hyperref}
\usepackage{amsmath}
\usepackage[capitalize,noabbrev]{cleveref}
\usepackage{pifont}

\iclrfinalcopy
\usepackage{xcolor}

\newcommand{\mycomment}[1]{}
\usepackage{amssymb}
\usepackage{mathtools}
\usepackage{amsthm}
\usepackage{placeins}[section]
\usepackage{todonotes}

\usepackage{xspace}

\newcommand{\task}{\mathsf{ImageNet{\text -}UA}}
\newcommand{\cifartask}{\mathsf{CIFAR{\text -}10{\text -}UA}}
\newcommand{\cifar}{\mathsf{CIFAR{\text -}10}}
\newcommand{\cifarfive}{\mathsf{CIFAR{\text -}10{\text -}50M}}

\newcommand{\UAA}{\mathsf{UA2}}
\newcommand{\eps}{\varepsilon}
\newcommand{\JPEG}{\mathsf{JPEG}}

\theoremstyle{plain}

\usepackage{hyperref}
\usepackage{url}

\title{Testing Robustness Against Unforeseen \\ Adversaries}

\newcommand{\eqcontrontrib}{\protect\thanks{Equal contribution.}}
\author{%
  Maximilian Kaufmann\eqcontrontrib \\
  UK Frontier AI Taskforce\\
  \And 
  Daniel Kang$^*$ \\
  UC Berkeley \\
  \And
  Yi Sun$^*$  \\
  University of Chicago \\
  \And
  Steven Basart \\
  Center for AI Safety \\
  \And
  Xuwang Yin \\
  Center for AI Safety \\
  \And
  Mantas Mazeika \\ Center for AI Safety \\
  \And
  Akul Arora\\
  UC Berkeley \\
  \And
  Adam Dziedzic\\Toronto University \\
  \And
  Franziska Boenisch\\Vector Institute \\
  \And
  Jacob Steinhardt\\UC Berkeley \\
  \And
  Dan Hendrycks \\ Center for AI Safety \\
}

\begin{document}

\maketitle
\input{tex/abstract.tex}
\input{tex/1-introduction.tex}
\input{tex/2-related-work.tex}

\input{tex/3-unforseen-adversaries.tex}

\input{tex/4-imagenetua.tex}
\input{tex/5-insights.tex}

\input{tex/6-conclusion.tex}

\bibliography{main}
\bibliographystyle{includes/iclr2024_conference}

\appendix

\input{tex/appendix.tex}
\end{document}

%% file: tex/abstract.tex
\begin{abstract}
Adversarial robustness research primarily focuses on $L_p$ perturbations, and most defenses are developed with identical training-time and test-time adversaries. However, in real-world applications developers are unlikely to have access to the full range of attacks or corruptions their system will face. Furthermore, worst-case inputs are likely to be diverse and need not be constrained to the $L_p$ ball. To narrow in on this discrepancy between research and reality we introduce $\mathsf{ImageNet{\text -}UA}$, a framework for evaluating model robustness against a range of \emph{unforeseen adversaries}, including eighteen new non-$L_p$ attacks. To perform well on $\mathsf{ImageNet{\text -}UA}$, defenses must overcome a generalization gap and be robust to a diverse attacks not encountered during training. In extensive experiments, we find that existing robustness measures do not capture unforeseen robustness, that standard robustness techniques are beat by alternative training strategies, and that novel methods can improve unforeseen robustness. We present $\mathsf{ImageNet{\text -}UA}$ as a useful tool for the community for improving the worst-case behavior of machine learning systems.
\end{abstract}

%% file: tex/1-introduction.tex
\section{Introduction}
\label{sec:intro}

Neural networks perform well on a variety of tasks, yet can be consistently fooled by minor adversarial distortions~\citep{intruiging-properties, goodfellowexplaining}. This has led to an extensive and active area of research, mainly focused on the threat model of an ``$L_p$-bounded adversary'' that adds imperceptible distortions to model inputs to cause misclassification. However, this classic threat model may fail to fully capture many real-world concerns regarding worst-case robustness \citep{Gilmer2018MotivatingTR}. Firstly, real-world worst-case distributions are likely to be varied, and are unlikely to be constrained to the $L_p$ ball. Secondly, developers will not have access to the worst-case inputs to which their systems will be exposed to. For example, online advertisers use perturbed pixels in ads to defeat ad blockers trained only on the previous generation of ads in an ever-escalating arms race~\citep{tramer2018adversarial}. Furthermore, although research has shown that adversarial training can lead to overfitting, wherein robustness against one particular adversary does not generalize ~\citep{dai2022formulating,yu2021understanding,stutz2020confidence,tramer2019adversarial}, the existing literature is still focuses on defenses that train against the test-time attacks. This robustness to a train-test distribution shift has been studied when considering average-case corruptions \citep{hendrycks2018benchmarking}, but we take this to the worst-case setting.

We address the limitations of current adversarial robustness evaluations by providing a repository of nineteen gradient-based attacks, which are used to create $\task$\footnote{Code available at \href{https://github.com/centerforaisafety/adversarial-corruptions}{github.com/centerforaisafety/adversarial-corruptions}}---a benchmark for evaluating the \emph{unforeseen robustness} of models on the popular ImageNet dataset \citep{deng2009imagenet}. Defenses achieving high Unforeseen Adversarial Accuracy ($\UAA$) on $\task$ demonstrate the ability to generalize to a diverse set of adversaries not seen at train time, demonstrating a much more realistic threat model than the $L_p$ adversaries which are a focus of the literature.

\begin{figure*}[t]
\vspace{-0.2cm}
\centering
\includegraphics[width=0.93\linewidth]{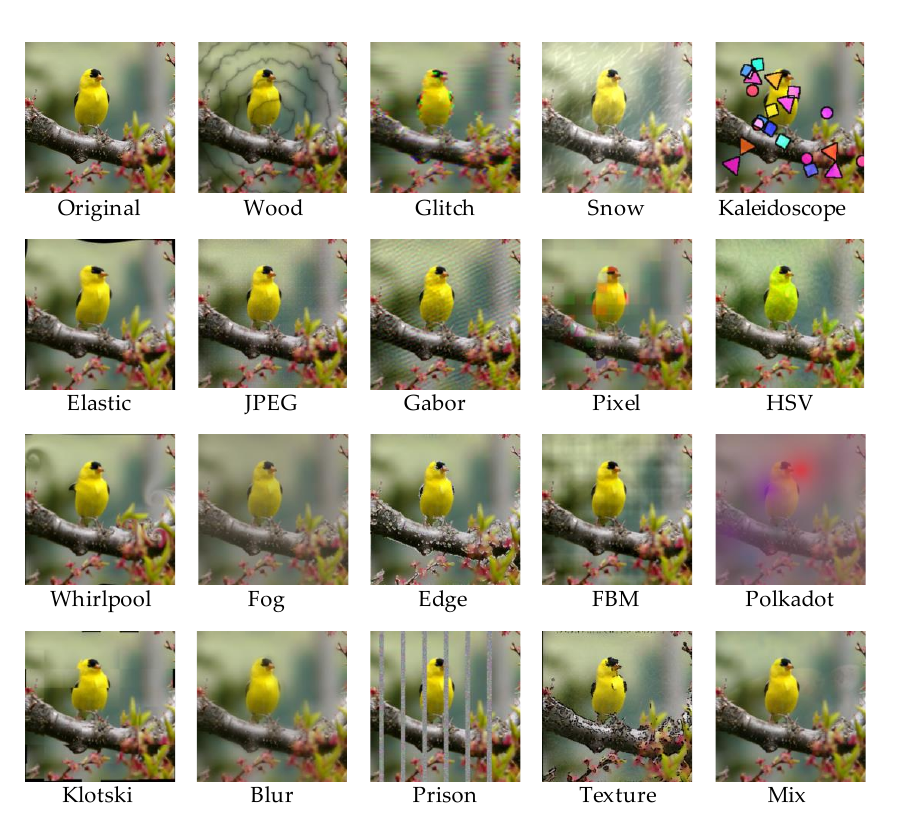}
\caption{\textbf{The full suite of attacks.} We present nineteen differentiable non-$L_p$ attacks as part of our codebase (eighteen of which are novel). For the purpose of visualization, higher distortion levels than are used in our benchmark have been chosen. See \cref{app:attack-images} for adversarial examples generated with the distortion levels used within our benchmark, and \cref{human-study} for a human study on semantic preservation.}
\label{fig:attack-images}

\end{figure*}

Our results show that unforeseen robustness is distinct from existing robustness metrics,  highlighting the need for a new measure which better captures the generalization of defense methods. We use $\task$  reveal that models with high $L_\infty$ attack robustness (the most ubiquitous measure of robustness in the literature) do not generalize well to new attacks, recommending $L_2$ as a stronger baseline. We further find that $L_p$  training can be improved on by alternative training processes, and suggest that the community focuses on methods with better generalization behavior. Interestingly, unlike in the $L_p$ case, we find that progress on CV benchmarks has at least partially tracked unforeseen robustness.  We are hopeful that $\task$ can provide an improved progress measure for defenses aiming to achieve  real-world worst-case robustness.

To summarize, we make the following contributions:

\begin{itemize}

\item We design eighteen novel non-$L_p$ attacks, constituting a large increase in the set of dataset-agnostic non-$L_p$ attacks available in the literature.
  
  \item We make use of these attacks to form a new benchmark ($\task$), standardizing and greatly expanding the scope of unforeseen robustness evaluation.
  
  \item We show that it $\UAA$  is distinct from existing robustness metrics in the literature, and demonstrates that classical $L_p$-training focused defense strategies can be improved on. We also measure the unforeseen robustness of a wide variety of techniques, finding promising research directions for generalizing adversarial robustness.

\end{itemize}

\begin{figure*}[ht!]
\vspace{-10pt}
\begin{center}
\includegraphics[width=1\linewidth]{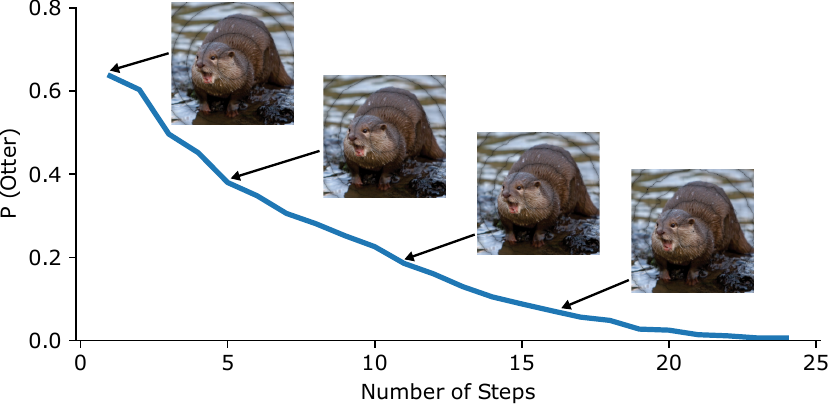}
\caption{\textbf{Progression of an attack.} As we optimize our differentiable corruptions, model performance decreases, while leaving the image semantics unchanged. Unoptimized versions of our attacks have a moderate impact on classifier performance, similar to common corruptions \citep{hendrycks2019benchmarking}, while optimized versions cause large drops in accuracy.
}
\label{fig:fourier-transform-perturbations}
\end{center}\vspace{-2mm}
\end{figure*}

%% file: tex/2-related-work.tex
\section{Related Work}\label{sec:related-work}\vspace{-0.2cm}
\textbf{Evaluating Adversarial Robustness.} Adversarial robustness is notoriously difficult to evaluate correctly
\citep{papernot2017practical,athalye2018obfuscated}. To this end, \citet{carlini2019evaluating} provide
extensive guidance for sound adversarial robustness evaluation.
Our $\task$ benchmark incorporates several of their recommendations, such as measuring attack success rates across several magnitudes of distortion and using
a broader threat model with diverse differentiable attacks. 
Existing measures of adversarial robustness  \citep{croce2020reliableautoattack,moosavi2015deepfool, weng2018evaluating} almost exclusively, apply only to attacks optimizing over an $L_p$-ball, limiting their applicability for modeling robustness to new deployment-time adversaries.

\textbf{Non-$L_p$ Attacks.} Many attacks either use generative models 
\citep{Song2018ConstructingUA,Qiu2019SemanticAdvGA} that are often hard to bound and are susceptible to instabilities, or make us of expensive brute-force search techniques \citet{engstrom2017rotation}.
We focus on attacks which
are fast by virtue of differentiability, applicable to  variety of datasets and independent of auxiliary generative models. Previous works presenting suitable attacks include \citet{laidlaw2019functional,shamsabadi2021semanticallycolorfilter,Zhao2019TowardsLY}, who all transform the underlying color space of an image and \citet{xiao2018spatially} who differentiably warp images, and which we adapt to create our own Elastic attack. The literature does not have a sufficiently diverse set of suitable adversaries to effectively test the generalization properties of defenses, causing us to develop out suite of attacks.

\textbf{Unforeseen and Multi-attack Robustness.} There exist  defense methods which seek to generalize across an adversarial train-test gap \citep{dai2022formulating,perceptualdistance, lin2020dual}.
Yet, comparison between these methods is challenging due to the lack of a standardized benchmark and an insufficient range of adversaries to test against. We fill this gap by implementing a unified benchmark for testing unforeseen robustness. The more developed field of multi-attack robustness \citep{tramer2019adversarial} aims to create models which are robust to a range of attacks, but works generally focus on a union of $L_p$ adversaries \citep{maini2020adversarial,madaan2021learning,croce2022adversarial} and do not enforce that test time adversaries have to differ from those used during training.\looseness=-1

\textbf{Common corruptions} Several of our attacks (Pixel, Snow, JPEG and Fog) were inspired by existing common corruptions \citep{hendrycks2018benchmarking}. We fundamentally change the generation methods to make these corruptions differentiable, allowing us to focus on worst-case robustness instead of the average-case robustness (see  \cref{sec:existing-robustness} for empirical an empirical comparison).

%% file: tex/3-unforseen-adversaries.tex
\section{The Unforeseen Robustness Threat Model}
 \begin{figure*}[t]
\includegraphics[width=\textwidth]{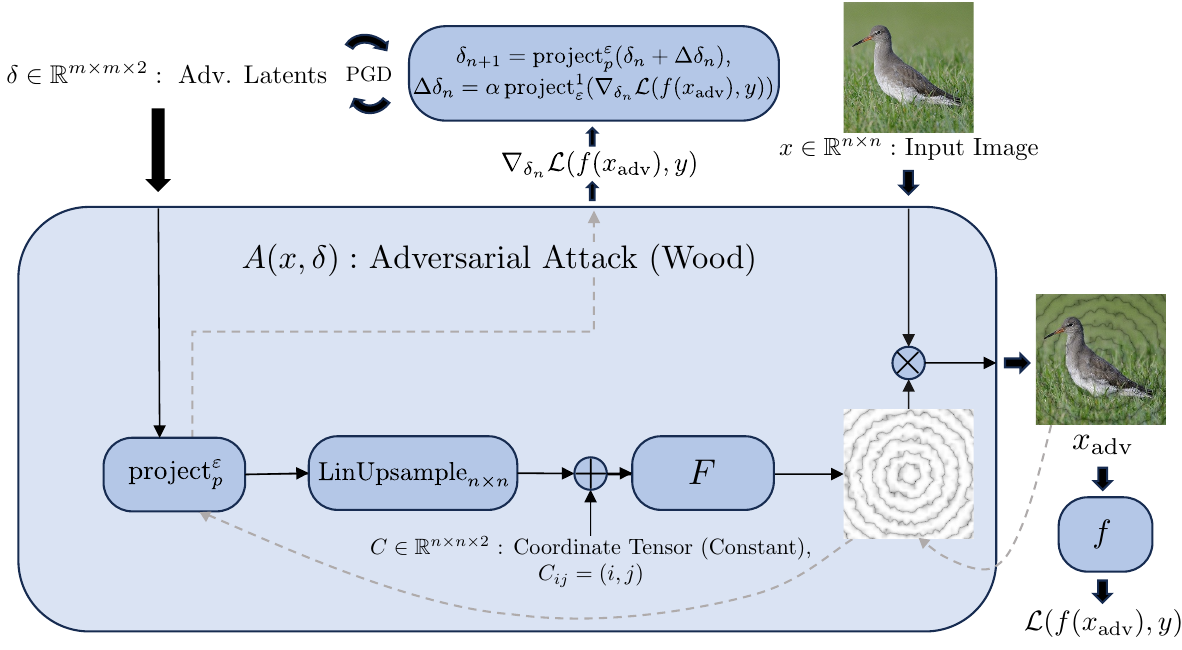}
 \caption{\textbf{An illustrative example of one of our attacks.} All of our attacks function by performing PGD optimization on a set of latent variables. In the case of the Wood attack, these latent variables are inputs to concentric sine waves ($F(x,y) = \sin(\sqrt{x^2 + y^2})$) which are overlaid on the image. We design effective attacks which are fast, easy to optimize, precisely bound, preserve image semantics, are portable across datasets and have variable intensity through the $\varepsilon$ parameter.}
\label{fig:wood-attack-explained}
 \end{figure*}

\textbf{Action Space of Adversaries.} The allowed action space of an adversary is defined using
a \emph{perturbation set} $S_x$ of potential adversarial examples for each input $x$. Given such a set, and a classifier $f$ which correctly classifies a point $x$ with its ground truth label $y$, an \emph{adversarial example} $x_\text{adv}$ is defined to be a member the perturbation set $S_x$ which causes the classifier to give an incorrect prediction: \begin{equation}\label{adv-examples-def} x_\text{adv} \in S_x : f(x_\text{adv})\neq f(x)\end{equation}
Then, under some distribution $\mathcal{D}$ of interest, the task of adversarial defenses is typically to achieve high accuracy in the face of an adversary which is allowed to optimse within the peturbation set.

The \emph{unforeseen robustness} of a classifier is the classifier's accuracy under some distribution of adversaries: 

\begin{equation}\label{adv-population-accuracy-defn}\mathbb{E}_{(x,y),A \sim \mathcal{D},\mathcal{A}}\left[\min_{x_\text{adv} \in S^A_x}\{\mathbf{1}_{f(x_{\text{adv}}) = y}\}\right].\end{equation}

This is similar to the usual \emph{adversarial accuracy} \citep{madry2017towards} , but instead of including  a single $L_p$ adversary, we define a diverse distribution of adversaries $\mathcal{A}$ (where each adversary $A \in \mathbf{Dom}(\mathcal{A})$ defines a different perturbation set $S^A_x$ for each input $x$).

\textbf{Information Available to the Adversaries.} To ensure that our adversaries are as strong as possible \citep{carlini2019evaluating}, and to avoid the usage of  expensive black-box optimization techniques, we allow full white-box access to the victim models.

\textbf{Constraints on the Defender.} We enforce that defenders allow adversaries to compute gradients, in line which previous work demonstrating that defenses relying on masking of gradients are ineffective \citep{athalye2018obfuscated}. We also enforce that defenses do not make use of access to adversaries which are part of the test-time distribution $\mathcal{A}$. This assumption of unforeseen adversaries is contrary to most of the literature where the most powerful defenses involve explicitly training against the test time adversaries \citep{towardsmadry}, and allows us to model more realistic real-world situations where it is unlikely that defenders will have full knowledge of the adversaries at deployment time.

%% file: tex/4-imagenetua.tex
\section{Measuring Unforeseen Robustness}
To evaluate the unforeseen robustness of models, we introduce a new evaluation framework consisting of a benchmark $\task$ and metric $\UAA$ (Unforeseen Adversarial Accuracy). We also further release our nineteen (eighteen of which novel) approaches for generating non-$L_p$ adversarial examples. We performed extensive sweeps to find the most effective hyperparameters for all of our attacks (see \cref{app:hyperparameters}).

\subsection{Generating adversarial examples}
\label{generation-strategy-attacks}
Our attacks use a unified generation strategy: Each of our adversaries is defined by a differentiable function $A$
, generating an adversarial input $x_\text{adv}$ from an input image $x$ and some latent variables $\delta$:
\begin{equation}
    x_\text{adv} = A(x,\delta) \text{.}
    \label{adv-generation-strategy}
\end{equation}

\begin{table}

\centering
                \begin{tabular}{lcc}\toprule
                Model & $L_{\infty}$ ($\varepsilon = 4/255$) & UA2 \\\midrule
                Dinov2 Vit-large  & 27.7 & \textbf{27.2} \\
                Convnext-V2-large IN-1k+22K  & 0.0 & 19.2 \\
                Swin-Large ImageNet1K & 0.0& 16.2 \\\midrule
                ConvNext-Base $L_{\infty}$, ($\varepsilon = 8/255$)  & \textbf{58.0} & 22.3 \\
                Resnet-50, $L_{\infty}$ ($\varepsilon = 8/255$) & 38.9 & 10 \\
                Resnet-50 $L_2$, ($\varepsilon = 5$)  & 34.1 & 13.9 \\ 
                \bottomrule  \\
                \end{tabular}
                \label{tab:lp-different}
\caption{\textbf{$L_p$ robustness is not necessary for unforeseen robustness.} We highlight some of the models which achieve high $\UAA$, while still being susceptible to $L_p$ attacks. These models demonstrate that unforeseen robustness is distinct from achieving $L_p$ robustness.}
            \end{table}

To control the strength of our adversary, we introduce an $L_p$ constraint to the variables $\delta$ (using $p=\infty$ or $p=2$ ).
We define our perturbation sets in terms of these allowed ranges of optimization variables, \textit{i.e.}, for attack $A$ with epsilon constraint $ \varepsilon$:
\[S^{A, \varepsilon}_x = \{ A(x,\delta) \mid \|\delta\|_p \leq  \varepsilon \}\label{unforeseen-adv-perturbation-set}.\]

As is typical in the literature \citep{towardsmadry}, we re-frame the finding of adversarial examples in our perturbation set \cref{unforeseen-adv-perturbation-set} as a continuous optimization problem, seeking $\delta_\text{adv}$ which solves:
\begin{equation}
\label{adv-optimisation-problem}
\delta_\text{adv} =\underset{\delta : \|\delta\|_p \leq  \varepsilon}{\operatorname{argmin}}\{\mathcal{L}(f(A(x,\delta)),y)\}\text{,}
\end{equation}
and we then use the popular method of Projected Gradient Descent (PGD) \citep{towardsmadry} to find an approximate solution to \cref{adv-optimisation-problem}.

\begin{figure}[t]
\centering
\subcaptionbox{\label{fig:num-steps-accuracy}Performance with increased optimization.}{
\includegraphics[width=0.48\linewidth]{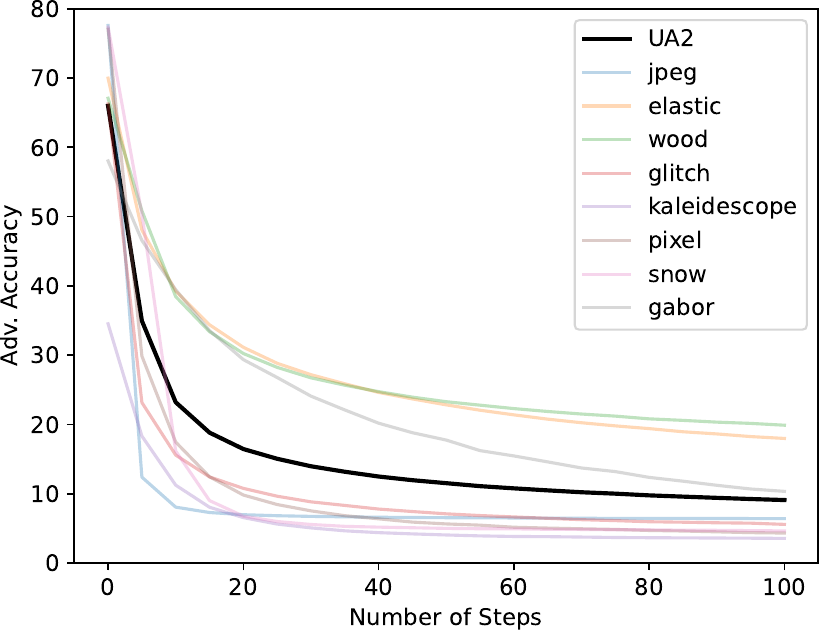}}
\subcaptionbox{\label{fig:distortion-level-accuracy}Performance as distortion size is varied}{
\includegraphics[width=0.48\linewidth]{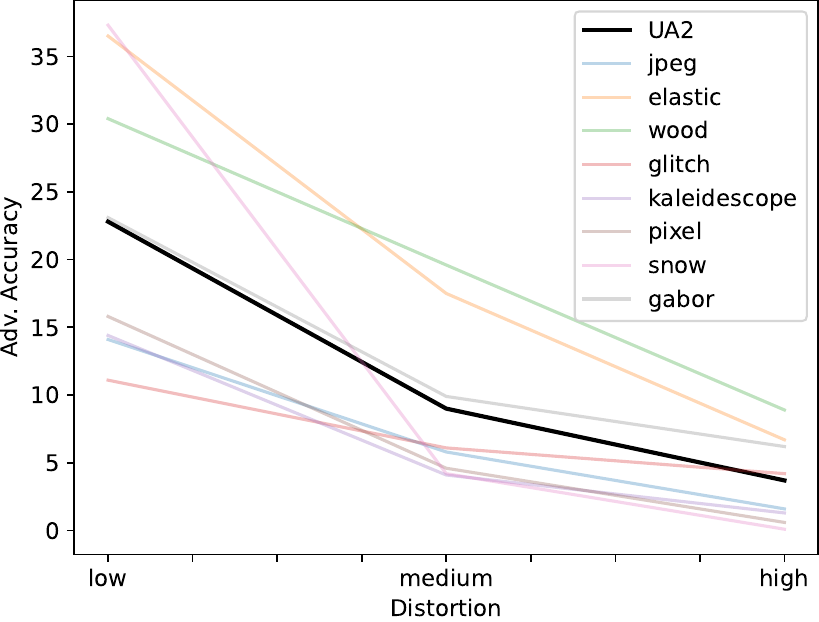}}
\caption{\textbf{Attack effectiveness increases with optimization pressure and distortion budget.} We average performance against our  core attacks across all our benchmarked models, demonstrating that our attacks respond to increased optimization pressure (\cref{fig:num-steps-accuracy}). We further demonstrate the importance of the gradient-based nature by comparing random grid search to our gradient-based method in \cref{app:grid-vs-gradient}. Furthermore, we demonstrate the ability for our attack stength to be customisable by showing that increasing distortion budget reduces model performance (\cref{fig:distortion-level-accuracy}).}
\end{figure}

This formulation helps us ensure that our attacks are independent of auxiliary generative models, add minimal overhead compared to the popular PGD adversary (see \cref{app:computation-time-attack}), are usable in a dataset-agnostic ``plug-and-play" manner, can be used with existing optimization algorithms (see \cref{fig:num-steps-accuracy} for behavior of attacks under optimization), come with a natural way of varying intensity through adjusting $\varepsilon$ parameter (see \cref{fig:distortion-level-accuracy} for behavior under varying $\varepsilon$), and have precisely defined perturbation sets which are not dependent on the solutions found to a relaxed constrained optimization problem. As discussed in \cref{sec:related-work}, this is not the case for most existing attacks in the literature, prompting us to design our new attacks.

\subsection{Core attacks}
\label{sec:eight-core-attacks}

To provide fast evaluation, we select eight core attacks to form the focus of our evaluation for unforeseen robustness.
We select the core set for diversity and effectiveness across model scale, leaving the other eleven attacks within our repository for the tuning of defense hyperperparameters and for a more complete evaluation of new techniques.  The eight core attacks are:

\noindent\textbf{Wood.} \quad The wood attack is described in \cref{fig:wood-attack-explained}.

\noindent\textbf{Glitch.}\quad Glitch simulates a common behavior in corrupted images of colored fuzziness. Glitch greys out the image, splitting it into horizontal bars, before independently shifting color channels within each of these bars.

\noindent\textbf{JPEG.} \quad The $\JPEG$ compression algorithm functions by encoding small image patches using the discrete cosine transform, and then quantizing the results. The attack functions by optimizing $L_\infty$-constrained perturbations within the  $\JPEG$-encoded space of compressed images and then 
reverse-transforming to obtain the image in pixel space, using ideas from \citet{shin2017jpeg} to make this differentiable.

\noindent\textbf{Gabor.} \quad Gabor spatially occludes the image with visually diverse Gabor noise \citep{lagae2009procedural}, optimizing the underlying sparse tensor which the Gabor kernels are applied to.

\noindent\textbf{Kaleidoscope.}\quad  Kaleidoscope overlays randomly colored polygons onto the image, and then optimizes both the homogeneous color of the inside of the shape, and the darkness/lightness of the individual pixels on the shape's border, up to an $L_\infty$ constraint.

\noindent\textbf{Pixel.}\quad Pixel modifies an image so it appears to be of lower quality, by first splitting the image into  $m \times m$ ``pixels'' and then and averaging the image color within each block. The optimization variables $\delta$ then control the level of pixelation, on a per-block bases.

\noindent\textbf{Elastic.}\quad Our only non-novel attack. Elastic is adapted from \citep{xiao2018spatially}, functioning by which warping the image  by distortions $x' = \mathrm{Flow}(x, V)$, where
$V: \{1, \ldots, 224\}^2 \to \mathcal{R}^2$ is a vector field on pixel space, and $\mathrm{Flow}$ sets
the value of pixel $(i, j)$ to the bilinearly interpolated original value at $(i, j) + V(i, j)$.
To make the attack
suitable for high-resolution images, we modify the original attack by passing a gaussian kernel over $V$ .

\noindent\textbf{Snow.} \quad Snow functions by optimising the intensity of individually snowflakes within an image, which are created by passing a convolutional filter over a sparsely populated tensor, and then optimising the non-zero entries in this tensor.

\label{sec:ua2-metric-definition}

\subsection{\texorpdfstring{$\task$}\space: a new benchmark for unforeseen robustness}
We introduce $\task$, a benchmark for evaluating the unforeseen robustness of image classifiers on the popular ImageNet dataset \citep{deng2009imagenet}. We also develop $\cifar$ equivalent $\cifartask$ for computationally efficient evaluation of defense strategies and attack methods.

The unforeseen robustness achieved by a defense is quantified using a new metric, Unforeseen Adversarial Accuracy ($\UAA$), which measures the robustness of a given classifier $f$ across a diverse range of unforeseen attacks. In line with \cref{adv-population-accuracy-defn} we model the deployment-time population of adversaries $\mathcal{A}$ as a categorical distribution over some finite set $\mathbf{A}$, with a distortion level $\epsilon_A$ for each adversary $A \in \mathbf{A}$. \cref{adv-population-accuracy-defn} then reduces to:

\[ \UAA :=\frac{1}{|\mathbf{A}|}\sum_{A \in \mathbf{A}}\ \operatorname{Acc}(A,\epsilon_A,f)\]
where $\operatorname{Acc}(A,\varepsilon_a,f)$ denotes the adversarial accuracy  of classifier $f$ against attack $A$ at distortion level $\eps_A$. We select the population of adversaries to be the eight core adversaries from \cref{sec:eight-core-attacks}, setting $\mathbf{A}$= \{JPEG, Elastic, Wood, Glitch, Kaleidoscope, Pixel, Snow, Gabor\}. 

We further divide our benchmark by picking three different distortion levels for each attack, leading to three different measures of unforeseen robustness: $\UAA_\text{low}$, $\UAA_\text{med}$ and $\UAA_\text{high}$ (see \cref{app:hyperparameters} for specific $\varepsilon$ values used within this work), and we focus on focus on $\UAA_{\text{med}}$ for all of our reports, referring to this distortion level as simply $\UAA$. As distortion levels increase, model performance decreases (\cref{fig:distortion-level-accuracy}). We perform a human study (\cref{human-study}) to ensure $\UAA_\text{med}$ preserves image semantics.

   \begin{table}[t!]

   \label{tab:pareto-frontier}   
   \resizebox{\textwidth}{!}{
    \setlength\tabcolsep{4pt}
    \begin{tabular}{lrrrrrrrrrrr}
    \toprule
     Model & Clean Acc. & $L_\infty$ & UA2 & JPEG & Elastic & Wood & Glitch & Kal. & Pixel & Snow & Gabor \\ \midrule
         DINOv2 ViT-large Patch14 & 86.1 & 15.3 & \bfseries 27.7 & 14.3 & \bfseries 42.6 & 39.7 & 17.7 & \bfseries 46.2 & \bfseries 17.2 & 14.2 & 29.9\\
          ConvNeXt-V2-large IN-1K+22K & \bfseries 87.3 & 0.0 & 19.2 & 0.0 & 39.1 & 34.4 & 21.4 & 16.1 &  15.5 & 4.0 & 23.1 \\
ConvNeXt-V2-huge IN-1K &  86.3 & 0.0 & 17.7 & 0.0& 42.5 & 21.2 & \bfseries 23.8 & 24.3 & 6.6 &  0.7 & 22.2 \\
        ConvNeXt-base, $L_\infty$ (4/255) & 76.1 & \bfseries 58.0 & 22.3 & 39.0 & 23.8 & \bfseries 47.9 & 12.9 & 2.5 & 9.7 & \bfseries 30.2 & 12.8 \\
    ViT-base Patch16, $L_\infty$ (4/255) & 76.8 & 57.1 & 25.8 & \bfseries 52.6 & 26.3 & 47.2 & 13.8 & 8.1 & 11.9 & 27.1 & 19.5 \\
    Swin-base IN-1K & 85.3 & 0.0 & 15.2 & 0.0 & 31.4 & 24.6 & 16.2 & 6.0 & 6.9 & 4.3 & \bfseries 32.0 \\
    \midrule
    ResNet-50 & 76.1 & 0.0 & 1.6 & 0.0 & 4.4 & 6.3 & 0.4 &  0.0 & 0.3 & 0.1 & 0.9\\
            ResNet-50 + CutMix & 78.6 & 0.5 & 6.1 & 0.2 & 17.9 & 15.5 & 2.5 & 0.1 &6.7 & 3.0 & 2.7\\
    ResNet-50, $L_\infty$ (8/255) & 54.5& 38.9 & 10.0 & 6.9 & 11.8 & 23.9 & 14.4 & 0.7 & 5.2 & 15.6 & 1.2\\
    ResNet-50, $L_2$ (5) & 56.1 & 34.1 & 13.9 & 39.7 & 11.9 & 19.4 & 12.2 & 0.3 &9.7 & 15.4 & 2.5\\
    \bottomrule\\
    \end{tabular}
    }
       \caption{\textbf{$\task$ baselines} We plot a range of models on the Pareto frontier on $\task$, as well as several baseline ResNet-50 models to compare between. We see a variety of techniques achieving high levels of robustness, demonstrating a rich space of possible interventions. The $L_\infty$ column tracks robustness against a PGD $L_\infty$ adversary with $\varepsilon = 4/255$. Numbers denote percentages.}
    \end{table}

%% file: tex/5-insights.tex
\begin{table}[!t]

\begin{minipage}{.41\linewidth}\hspace{0.3cm}
\begin{small}

\begin{tabular}{ccccc}\toprule
Training &  Train $\varepsilon $& Clean Acc. &UA2
 \\\midrule
 Standard & - &\textbf{76.1} &1.6\\\midrule
 \multirow{3}{*}{$L_2$} 
 &$   1$ &69.1 &6.4 \\
 &$ 3$ &62.8 &12.2  \\
 &$5$ &56.1  &\textbf{13.9}  \\\midrule
 \multirow{3}{*}{$L_\infty$} & $2/255$ &69.1  &6.4  \\
& $  4/255$ &63.9  &7.9  \\
&$  8/255$ &54.5  &10.0  \\
\bottomrule
\label{tab:adv-trained-results}

\end{tabular}
\caption{\textbf{$L_p$ training.} We train a range of ResNet-50 models against $L_p$ adversaries on $\task$}
\end{small}
\end{minipage}\hspace{0.34cm}
\begin{minipage}{.55\linewidth}
\begin{small}

\begin{tabular}{cccc}\toprule
Dataset & Training  & Clean Acc. &UA2 \\\midrule
\multirow{2}{*}{$\cifar$}
& $L_2,\varepsilon=1$ & 82.3 & 45.8\\
& $L_\infty, \varepsilon = 8/255 $ & 86.1 & 41.5 \\\midrule
\multirow{2}{*}{$\cifarfive$} 
& $L_2, \varepsilon = 0.5$ &95.2 &\textbf{51.2} \\
& $L_\infty, \varepsilon =4/255$ & 92.4 &\textbf{51.5} \\
\bottomrule
\end{tabular}
\label{tab:generated-diffusion-data}
\end{small}
\caption{\textbf{$L_p$ training on generated data.} We see the effect of training when training WRN-28-10 networks on $\cifarfive$, a 1000x larger diffusion-model generated version of $\cifar$ \citep{wang2023betterdiffusion}}
\end{minipage}
\end{table}

\section{Benchmarking for Unforeseen Adversarial Robustness}
\label{sec:results}
In this section, we evaluate a range of models on our standardized benchmarks $\task$ and $\cifartask$. We aim to present a set of directions for future work, by comparing a wide range of methods. We also hope to explore how the problem of unforeseen robustness different from existing robustness metrics.

\newlength{\mylen}
\settowidth{\mylen}{Training Strategy}

\subsection{How do existing robustness measures relate to unforeseen robustness?}
\label{sec:existing-robustness}

We find the difference between existing popular metrics and $\UAA$, highlighting the differential progress made possible by  $\UAA$:

\textbf{Worst-case and average-case robustness behave differently.} We compare $\UAA$ to the average-case robustness metric given by ImageNet-C \citep{hendrycks2019benchmarking}.  As shown in \cref{app:imagenet-c-unforeseen}, we find that performance on this benchmark correlates with non-optimized versions of our attacks. However, the optimised versions of our attacks have model robustness profiles more similar to $L_p$ adversaries. We see believe this is as after optimisation $\UAA$ becomes a measure of worst case robustness, similar to $L_p$ robust accuracy--- contrasting with the average-case robustness considered in ImageNet-C.

\textbf{$L_p$ robustness is correlated, but distinct, from unforeseen robustness.} As shown  in \cref{app:lp-correlation}, unforeseen robustness is correlated with $L_p$ robustness. Our attacks  also show similar properties to $L_p$ counterparts, such as the ability for black-box transfer (\cref{app:black-box-transfer}). However,  many models show susceptibility to  $L_p$ adversaries while still performing well on $\UAA$ (\cref{tab:lp-different}), and a range of strategies beat $L_p$ training baselines \cref{sec:improving-unforeseen-robustness} . We conclude that $\UAA$ is distinct from $L_p$ robustness, and present $\UAA$ as an improved progress measure when working towards real-world worst-case robustness.

\textbf{$L_2$-based adversarial training outperforms $L_\infty$} We see that $L_p$ adversarial training increases the unforeseen robustness of tested models, with $L_2$ adversarial training providing the largest increase in $\UAA$ over standard training (1.6\% $\to$ 13.9\%), beating models which are trained against $L_\infty$ adversaries (1.6\% $\to$ 10.0\%). We present $L_2$ trained models as a strong baseline for unforeseen robustness, noting that the discrepancy between $L_\infty$ and $L_2$ training is particularly relevant as $L_\infty$ is the most ubiquitous measure of adversarial robustness in the literature. 

\begin{table}[t]
\vspace{-10pt}

\label{tab:pixmix-adv2}
\centering
\begin{tabular}{lrcr} %
\toprule
Training Strategy & Train $\varepsilon$ & Clean Acc. & UA2 \\
\midrule
PixMix & - & \textbf{95.1} & 15.00 \\ 
\midrule
$L_\infty$ & 4/255 & 89.3 & 37.3 \\
$L_\infty$ + PixMix & 4/255 & 91.4 & \textbf{45.1} \\ 
\midrule
$L_\infty$ & 8/255 & 84.3 & 41.4 \\
$L_\infty$ + PixMix & 8/255 & 87.1 & \textbf{47.4}\\
\bottomrule
\end{tabular}
\caption{\textbf{PixMix and $L_p$ training.} We compare UA2 performance on CIFAR-10 of models trained with PixMix and adversarial training. Combining PixMix with adversarial training results in large improvements in $\UAA$, demonstrating the potential for novel methods to improve UA2. All numbers denote percentages, and $L_\infty$ training was performed with the TRADES algorithm.}
\end{table}

\subsection{How can we improve Unforeseen Robustness?} \label{sec:improving-unforeseen-robustness}
We find several promising directions that improve over $L_p$ training, and suggest that the community should focus more on techniques which we demonstrate to have better generalization properties:

\textbf{Combining image augmentations and $L_\infty$ training.} We combine PixMix and $L_\infty$ training, finding that this greatly improves unforeseen robustness over either approach alone ($37.3 \to 45.1$, see \Cref{tab:pixmix-adv2}). This is a novel training strategy which beats strong baselines by combining two distinct robustness techniques ($L_p$ adversarial training and data augmentation). The surprising effectiveness of this simple method suggests that unforeseen robustness may foster the development of new methods.

\textbf{Multi-attack robustness.} To evaluate how existing work on robustness to a union of $L_p$ balls may improve unforeseen robustness, we use $\cifartask$ to evaluate a strong multi-attack robustness baseline by \citep{madaan2021learninggeneratenoise}, which trains a Meta Noise Generator (MNG) that learns the optimal training perturbations to achieve robustness to a union of  $L_p$ adversaries. For WRN-28-10 models on $\cifartask$, we see a large increase in unforeseen robustness compared to the best $L_p$ baseline ($21.4\% \to 51.1\%$, full results in \cref{app:non-lp-training} ), leaving scaling of such methods to full $\task$ for future work.

\textbf{Bounding perturbations with perceptual distance.} We evaluate the $\UAA$ of models trained with Perceptual Adversarial Training (PAT) \citep{perceptualdistance}. PAT functions by training a model against an adversary bounded by an estimate of the human perceptual distance, computing the estimate by using the hidden states of an image classifier. For computational reasons we train and evaluate ResNet-50s on a 100-image subset of $\task$, where this technique outperforms the best $L_p$ trained baselines ($22.6 \to 26.2$, full results in \cref{app:non-lp-training}).

\textbf{Regularizing high-level features.} We evaluate Variational Regularization (VR) \citep{dai2022formulating}, which adds a penalty term to the loss function for variance in higher level features. We find that the largest gains in unforeseen robustness come from combining VR with PAT, improving over standard PAT ($26.2 \to 29.5$, on a 100 class subset of $\task$, full results in \cref{app:non-lp-training}).

\begin{table}[!t]
\vspace{-10pt}
\label{fig:combined_tables_cv_robustness}
\begin{minipage}{.45\linewidth}
\centering

\begin{tabular}{lrr}\toprule
Training &Clean Acc. &UA2 \\\midrule
Standard &76.1 &1.0 \\
Moex &\textbf{79.1} &\textbf{6.0}  \\
CutMix &78.6 &\textbf{6.0} \\
Deepaugment + Augmix &75.8 &1.8 \\
\bottomrule
\label{tab:data-augmentation-robustness}
\end{tabular}
\caption{\textbf{Effects of data augmentation on $\UAA$.} We evaluate the $\UAA$ of a range of data-augmented ResNet50 models.}
\end{minipage}
\hfill
\begin{minipage}{.45\linewidth}
\centering

\begin{tabular}{lcc}\toprule
Model &Clean Acc. & UA2 \\\midrule
ConvNeXt-V2-28.6M &\textbf{83.0} &\textbf{9.8} \\
ConvNeXt-V1-28M &82.1 &5.1 \\
\midrule
ConvNeXt-V2-89M &\textbf{84.9} &\textbf{14.9} \\
ConvNeXt-V1-89M &83.8  &9.7  \\\midrule
ConvNeXt-V2-198M &\textbf{85.8} &\textbf{19.1} \\
ConvNeXt-V1-198M &84.3  &10.6 \\\midrule
\label{tab:pretraining-comparison}
\end{tabular}

\caption{\textbf{Effects of pretraining and regularization on $\UAA$.} We compare the unforeseen robustness of ConvNext-V1 and ConvNext-V2 models of equivalent sizes, finding ConvNext-V2 models improve over their V1 versions. }

\end{minipage}
\centering

\end{table}

\subsection{How has progress on CV benchmarks tracked unforeseen robustness?}

\textbf{Computer vision progress has partially tracked unforeseen robustness.} Comparing the $\UAA$ of ResNet-50 to ConvNeXt-V2-huge ($ 1\% \to 19.1\% \hspace{1mm}\UAA $) demonstrates the effects of almost a decade of CV advances, including self-supervised pretraining, hardware improvements, data augmentations, architectural changes and new regularization techniques. More generally, we find a range of modern architectures and training strategies doing well (see \cref{tab:pareto-frontier}, full results in \cref{fig:UA2-imagenet-medium}). This is gives a positive view of how progress on standard CV benchmarks has tracked underlying robustness metrics, contrasting with classical $L_p$ adversarial robustness where standard training techniques have little effect \citep{madry2017towards}.

\textbf{Scale, data augmentation and pretraining 
 improve robustness.} We do a more careful analysis of  how three of the most effective CV techniques have improved robustness.  As shown in \cref{tab:data-augmentation-robustness}, we find that data augmentation improves on unforeseen robustness, even in cases where they reduce standard accuracy. We  also compare the performance  of ConvNeXt-V1 and ConvNeXt-V2 models, which differ through the introduction of self-supervised pretraining and a new normalization layer. When controlling for model capacity these methods demonstrate large increase unforeseen robustness (see \cref{tab:pretraining-comparison}). We also note that DINOv2, the best performing model on our benchmark, is the product of self-supervised pretraining at a large scale.

%% file: tex/6-conclusion.tex
\FloatBarrier
\section{Conclusion}
In this paper, we introduced a new benchmark for
\emph{unforeseen adversaries} ($\task$) laying groundwork for future research in improving real world adversarial robustness.
We provide nineteen (eighteen novel) non-$L_p$ attacks as part of our repository, using these to construct a new metric $\UAA$ (Unforeseen Adversarial Accuracy). We then make use use this standardized benchmark to evaluate classical $L_p$ training techniques, showing
 that the common practice of $L_\infty$ training and evaluation may be misleading,
as $L_2$ shows higher unforeseen robustness. We additionally demonstrate that a variety of interventions outside of $L_p$ adversarial training can improve unforeseen robustness, both through existing techniques in the CV literature and through specialised training strategies. We hope that the $\task$ robustness framework will help guide adversarial robustness research, such that we continue making meaningful progress towards making machine learning safer for use in real-world systems.

%% file: tex/appendix.tex
\section{Hyperparameters}
\label{app:hyperparameters}
\subsection{Trained models}
To run our evaluations, we train a range of our own models to benchmark with:

\begin{itemize}
    \item CIFAR-10 WRN-28-10 robust models and TRADES models are respectively trained with the official code of ~\cite{rice2020overfitting} and ~\cite{zhang2019theoretically} with the default hyperparameters settings
    \item The PAT-VR models on ImageNet100 were trained using the official code from ~\cite{dai2022formulating} and employed the hyperparameter settings outlined in the code of~\cite{perceptualdistance}.
    \item ImageNet100 DINOv2~\cite{oquab2023dinov2} models are trained by finetuning a linear classification head on the ImageNet100 dataset. We used a SGD optimizer with learning rate of 0.001 and employed early-stopping.
\end{itemize}

\subsection{Model Reference}
We use a range of baseline models provided by other works, with model weights available as part of their open source distribution:
\begin{itemize}
\item \textbf{ImageNet} 
    \begin{itemize}
        \item ConvNeXt models are from \cite{liu2022convnet}
        \item ConvNeXt-V2 models are from \cite{woo2023convnext}
        \item ViT models are from \cite{steiner2022how}
        \item Swin models are from \cite{liu2021swin}
        \item Reversible-ViT models are from \cite{mangalam2022reversible}
        \item CLIP (ViT-L/14) is from \cite{radford2021learning}
        \item DINOv2 models are from \cite{oquab2023dinov2}
        \item MAE models are from \cite{he2022masked}
    \end{itemize}
\item \textbf{CIFAR-10}
    \begin{itemize}
        \item WideResNet TRADES models are from~\cite{zhang2019theoretically}
        \item WRN + Diffusion models are from~\cite{wang2023betterdiffusion}
        \item Meta noise models are from~\cite{madaan2021learninggeneratenoise}
        \item ResNet50 VR models are from~\cite{dai2022formulating}
        \item ReColorAdv models are from~\cite{laidlaw2019functional}
        \item StAdv modesl are from~\cite{xiao2018spatially}
        \item Multi attack models are from~\cite{tramer2018adversarial}
        \item The Multi steepest descent model is from~\cite{maini2020adversarial}
        \item PAT models are from~\cite{perceptualdistance}
        \item Pre-trained ResNet18 $L_\infty$, $L_2$ and $L_1$ models are from \cite{croce2022adversarial}
    \end{itemize}
\item \textbf{ImageNet100}
    \begin{itemize}
        \item ResNet50 PAT models are from \cite{perceptualdistance}
        \item ResNet50 PAT + VR models are from \cite{dai2022formulating}
        \item DINOv2 models are from \cite{oquab2023dinov2}
    \end{itemize}
\end{itemize}

\subsection{Attack Parameters}
To ensure that our attacks are maximally effective, we perform extensive hyper-parameter sweeps to find the most effective step sizes.
\label{sec:attack-hyperparameters}
\begin{table}[h!]
\centering
\label{tab:hyperparams-for-imagenet-attacks}
\resizebox{\textwidth}{!}{%
\begin{tabular}{@{}llllllll@{}}
\toprule
                        &  & Step Size & Num Steps & Low Distortion & Medium Distortion & High Distortion & Distance Metric \\ \midrule
\multirow{9}{*}{Core Attacks}
 & PGD  & 0.004    & 50       & 2/255           & 4/255              & 8/255            &      $L_\infty$           \\
 & Gabor  & 0.0025    & 100       & 0.02           & 0.04              & 0.06            &      $L_\infty$           \\
 & Snow         & 0.1    & 100 & 10    & 15    & 25    & $L_2$ \\ 
 & Pixel        & 1      & 100 & 3     & 5     & 10    & $L_2$ \\ 
 & JPEG         & 0.0024 & 80  & 1/255 & 3/255 & 6/255 &  $L_\infty$\\ 
 & Elastic      & 0.003  & 100 & 0.1   & 0.25  & 0.5   &  $L_2$\\ 
 & Wood         & 0.005  & 80  & 0.03  & 0.05  & 0.1   & $L_\infty$ \\ 
 & Glitch       & 0.005  & 90  & 0.03  & 0.05  & 0.07  & $L_\infty$ \\ 
 & Kaleidoscope & 0.005  & 90  & 0.05  & 0.1   & 0.15  &  $L_\infty$ \\  \midrule
\multirow{11}{*}{Extra Attacks} 
 & Edge      & 0.02       & 60         & 0.03           & 0.1               & 0.3             &        $L_\infty$         \\
 & FBM       & 0.006      & 30         & 0.03           & 0.06              & 0.3             &        $L_\infty$         \\
 & Fog       & 0.05       & 80         & 0.3            & 0.5               & 0.7             &        $L_\infty$         \\
 & HSV       & 0.012      & 50         & 0.01           & 0.03              & 0.05            &        $L_\infty$         \\
 & Klotski   & 0.01       & 50         & 0.03           & 0.1               & 0.2             &        $L_\infty$         \\
 & Mix       & 1.0        & 70         & 5              & 10                & 40              &        $L_2$           \\
 & Pokadot   & 0.3        & 70         & 1              & 3                 & 5               &        $L_2$           \\
 & Prison    & 0.0015     & 30         & 0.01           & 0.03              & 0.1             &        $L_\infty$         \\
 & Blur      & 0.03       & 40         & 0.1            & 0.3               & 0.6             &        $L_\infty$         \\
 & Texture   & 0.00075    & 80         & 0.01           & 0.03              & 0.2             &        $L_\infty$         \\
 & Whirlpool & 4.0        & 40         & 10             & 40                & 100             &        $L_2$           \\
 \bottomrule

\end{tabular}%
}
\caption{Attack parameters for $\task$ }
\end{table}

\begin{table}[!ht]
\centering

\label{tab:hyperparams-for-cifar-attacks}
\resizebox{\textwidth}{!}{%
\begin{tabular}{@{}llllllll@{}}
\toprule
                        &  & Step Size & Num Steps & Low Distortion & Medium Distortion & High Distortion & Distance Metric \\ \midrule
\multirow{9}{*}{Core Attacks}   
 & PGD  & 0.008    & 50       & 2/255           & 4/255              & 8/255            &      $L_\infty$           \\
& Gabor & 0.0025 & 80 & 0.02 & 0.03 & 0.04 & $L_\infty$ \\
& Snow & 0.2 & 20 & 3 & 4 & 5 & $L_2$ \\
& Pixel & 1.0 & 60 & 1 & 5 & 10 & $L_2$ \\
& JPEG & 0.0024 & 50 & 1/255 & 3/255 & 6/255 & $L_\infty$ \\
& Elastic & 0.006 & 30 & 0.1 & 0.25 & 0.5 & $L_2$ \\
& Wood & 0.000625 & 70 & 0.03 & 0.05 & 0.1 & $L_\infty$ \\
& Glitch & 0.0025 & 60 & 0.03 & 0.05 & 0.1 & $L_\infty$ \\
& Kaleidoscope & 0.005 & 30 & 0.05 & 0.1 & 0.15 & $L_\infty$ \\ 
 \midrule
\multirow{11}{*}{Extra Attacks} 
& Edge & 0.02 & 60 & 0.03 & 0.1 & 0.3 & $L_\infty$ \\
& FBM & 0.006 & 30 & 0.02 & 0.04 & 0.08 & $L_\infty$ \\
& Fog & 0.05 & 40 & 0.3 & 0.4 & 0.5 & $L_\infty$ \\
& HSV & 0.003 & 20 & 0.01 & 0.02 & 0.03 & $L_\infty$ \\
& Klotski & 0.005 & 50 & 0.03 & 0.05 & 0.1 & $L_\infty$ \\
& Mix & 0.5 & 30 & 1 & 5 & 10 & $L_2$ \\
& Pokadot & 0.3 & 40 & 1 & 2 & 3 & $L_2$ \\
& Prison & 0.0015 & 20 & 0.01 & 0.03 & 0.1 & $L_\infty$ \\
& Blur & 0.015 & 20 & 0.1 & 0.3 & 0.6 & $L_\infty$ \\
& Texture & 0.003 & 30 & 0.01 & 0.1 & 0.2 & $L_\infty$ \\
& Whirlpool & 16.0 & 50 & 20 & 100 & 200 & $L_2$ \\
 \bottomrule

\end{tabular}%
}
\caption{Attack parameters for $\cifartask$ }
\end{table}
\newpage
\FloatBarrier
\section{Descriptions of the 11 Additional Attacks.}
\label{app:full-attack-description}

\noindent\textbf{Blur.} \quad
Blur approximates real-world motion blur effects by passing a Gaussian filter over the original image and then does a pixel-wise linear interpolation between the blurred version and the original, with the optimisation variables controlling the level of interpolation. We also apply a Gaussian filter to the grid of optimisation variables, to enforce some continuity in the strength of the blur between adjacent pixels. This method is distinct from, but related to other blurring attacks in the literature \citep{NEURIPS2020watchoutblurring,guo2021learningtoadvblur}.

\noindent\textbf{Edge.} \quad
This attack functions by applying a Canny Edge Detector \citep{canny1986computational} over the image to locate pixels at the edge of objects, and then applies a standard PGD attack to the identified edge pixels.

\noindent\textbf{Fractional Brownian Motion (FBM).} \quad
FBM overlays several layers of Perlin noise \citep{perlin2005makingperlinnoise} at different frequencies, creating a distinctive noise pattern. The underlying gradient vectors which generate each instance of the Perlin noise are then optimised by the attack.

\noindent\textbf{Fog.} \quad
Fog simulates worst-case weather conditions, creating fog-like occlusions by adversarially optimizing parameters in the
diamond-square algorithm \citep{fournier1982computer} typically used to render stochastic fog effects.

\noindent\textbf{HSV.} \quad
 This attack transforms the image into the HSV color space, and then optimises PGD in that latent space. Due to improving optimisation properties, a gaussian filter is passed over the image.

\noindent\textbf{Klotski.} \quad
The Klotski attack works by splitting the image into blocks, and applying a differentiable translation to each block, which is then optimised.

\noindent\textbf{Mix.} \quad
The Mix attack functions by performing differntiable pixel-wise interpolation between the original image and an image of a different class. The level of interpolation at each pixel is optimised, and a gaussian filter is passed over the pixel interpolation matrix to ensure that the interpolation is locally smooth.

\noindent\textbf{Polkadot.} \quad
Polkadot randomly selects points on the image to be the centers of a randomly coloured circle, and then optimising the size of these circles in a differentiable manner. 

\noindent\textbf{Prison.} \quad
Prison places grey "prison bars" across the image, optimising only the images within the prison bars. This attack is inspired by previous ``patch'' attacks \citep{brown2017adversarial}, while ensuring that only the prison bars are optimised.

\noindent\textbf{Texture.} \quad
Texture works by removing texture information within an images, passing a Canny Edge Detector \citep{canny1986computational} over the image to find all the pixels which are at the edges of objects, and then filling these pixels in black---creating a silhouette of the original image. The other non-edge (or "texture") pixels are then whitened, losing the textural information of the image while preserving the shape. Per-pixel optimisation variable control the level of whitening.

\noindent\textbf{Whirlpool.} \quad
Whirlpool translates individual pixels in the image by a differentiable function creating a whirlpool-like warpings of the image, optimising the strength of each individual whirlpool.
\clearpage

\section{Attack Computation Time}
We investigate the execution times of our attacks, finding that most attacks are not significantly slower than an equivalent PGD adversary. 
\label{app:computation-time-attack}

\begin{figure}[h!]
\begin{center}
\includegraphics[width=\textwidth]{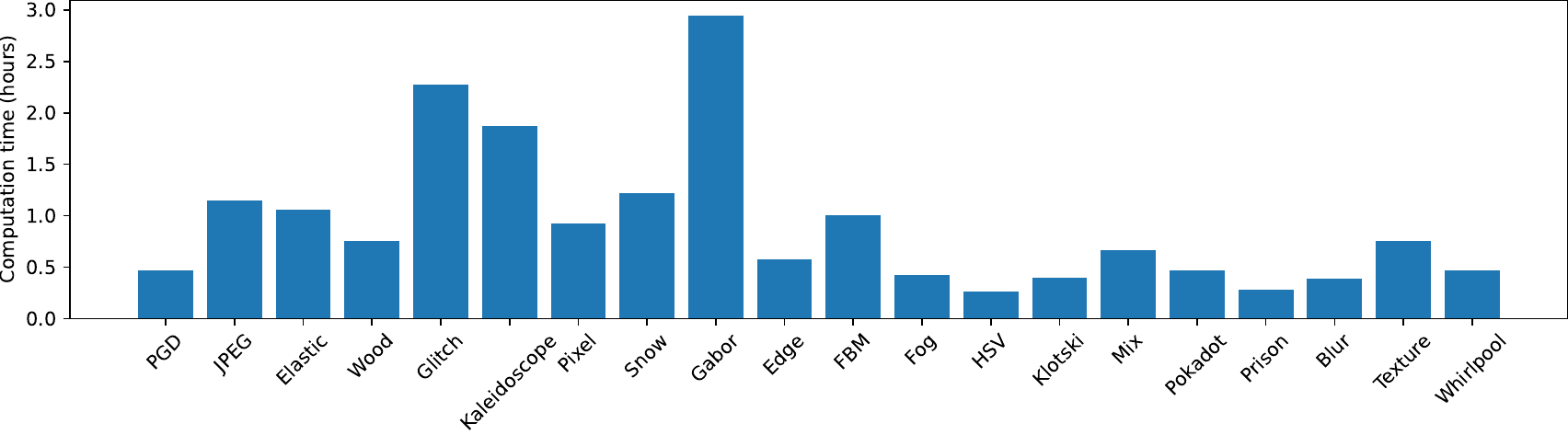}
\caption{Evaluation time of the attacks on the ImageNet test set using a ResNet50 model with batch size of 200 on a single A100-80GB GPU, Attack hyper-parameters are as described in \cref{app:hyperparameters}.}
\label{fig:computation-time}
\end{center}
\end{figure}
\newpage

\section{Full Results of Model Evaluations}
We benchmark a large variety of models on our dataset, finding a rich space of interventions affecting unforeseen robustness.
\label{scaling_all_attacks}
\subsection{ImageNet}

\begin{figure}[h!]
\begin{center}
\includegraphics[width=\textwidth]{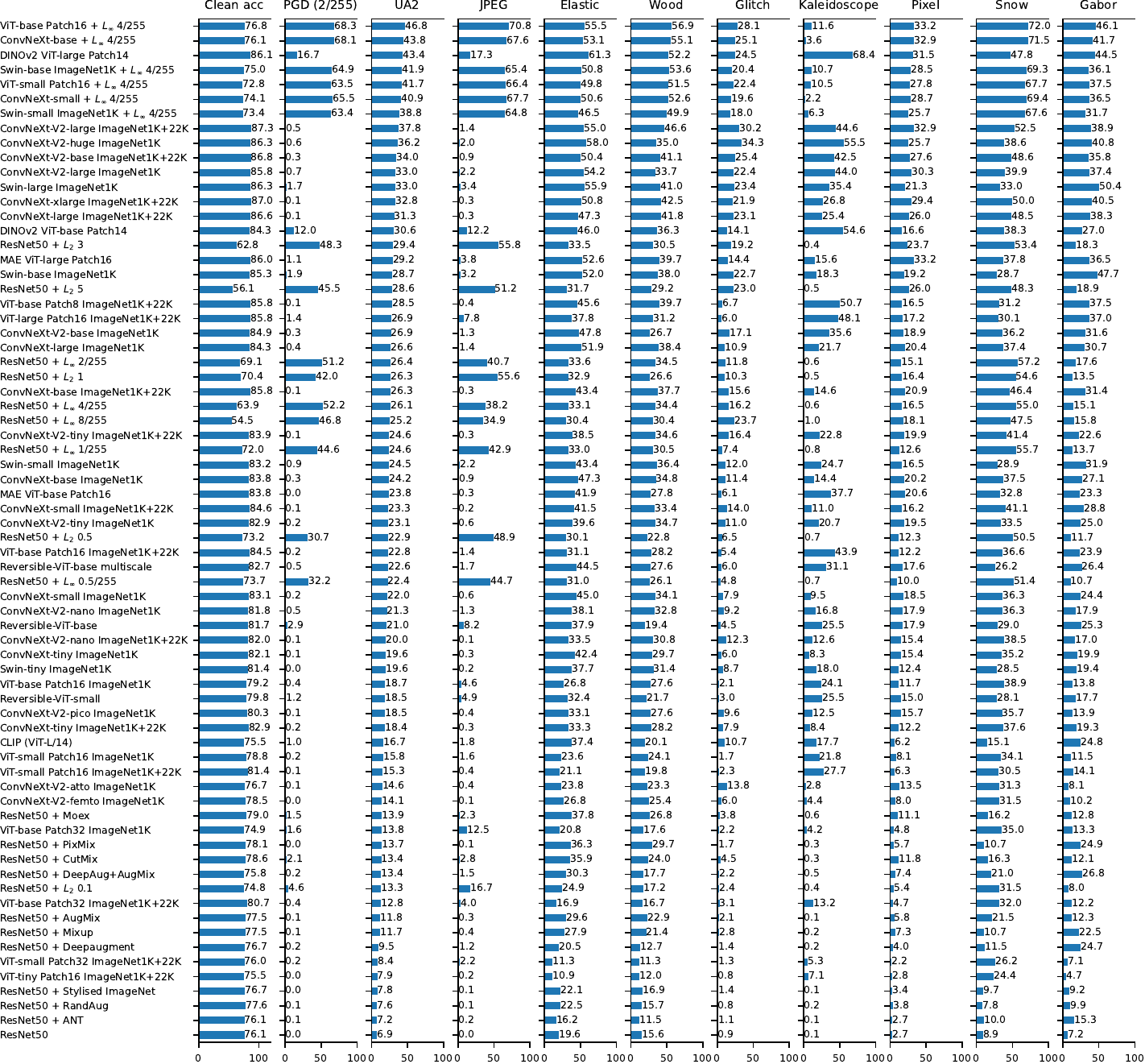}
\caption{ImageNet UA2 performance under low distortion.}
\label{fig:UA2-imagenet-low}
\end{center}
\end{figure}

\begin{figure}[h!]
\begin{center}
\includegraphics[width=\textwidth]{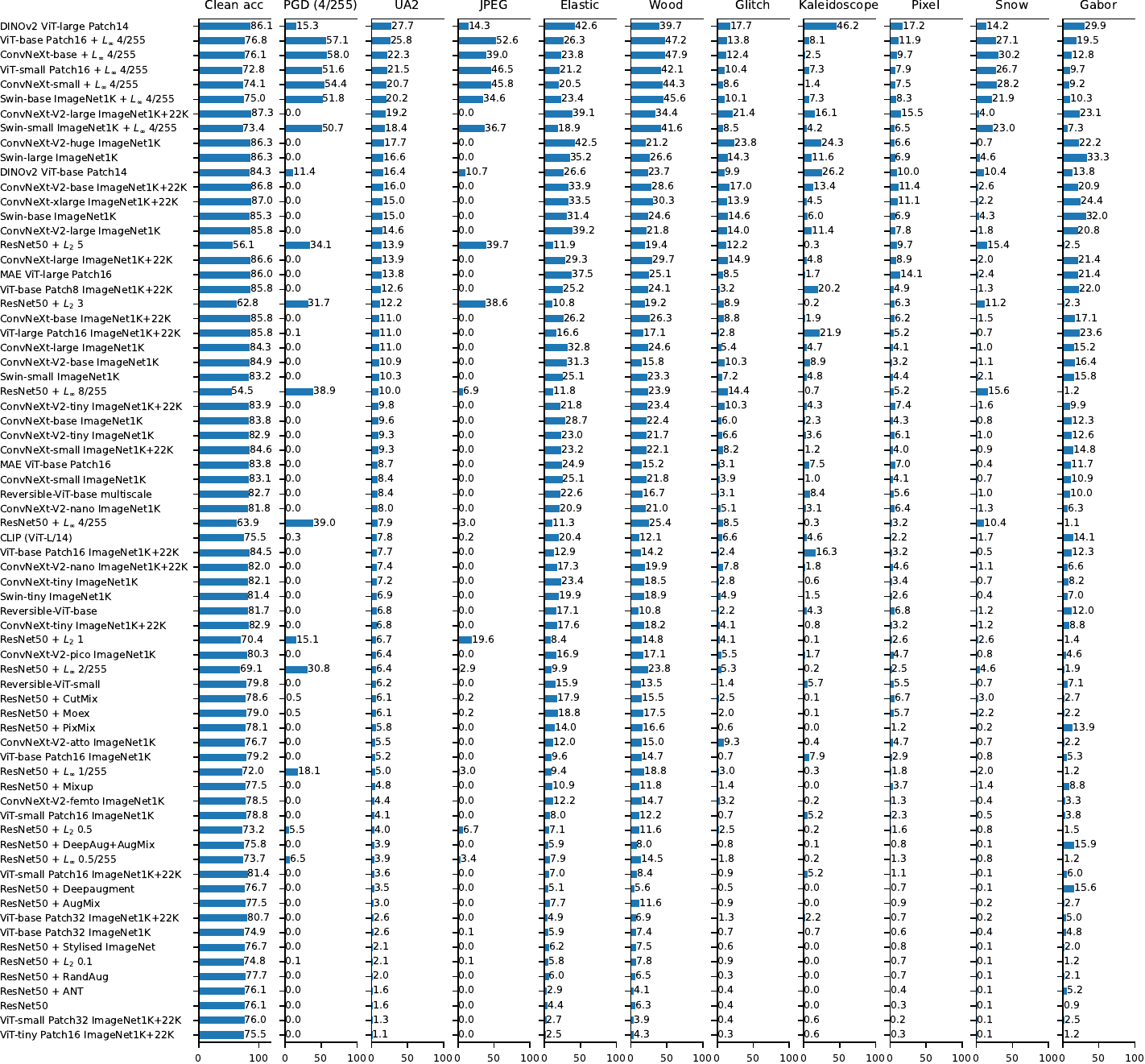}
\caption{ImageNet UA2 performance under medium distortion}
\label{fig:UA2-imagenet-medium}
\end{center}
\end{figure}

\begin{figure}[h!]
\begin{center}
\includegraphics[width=\textwidth]{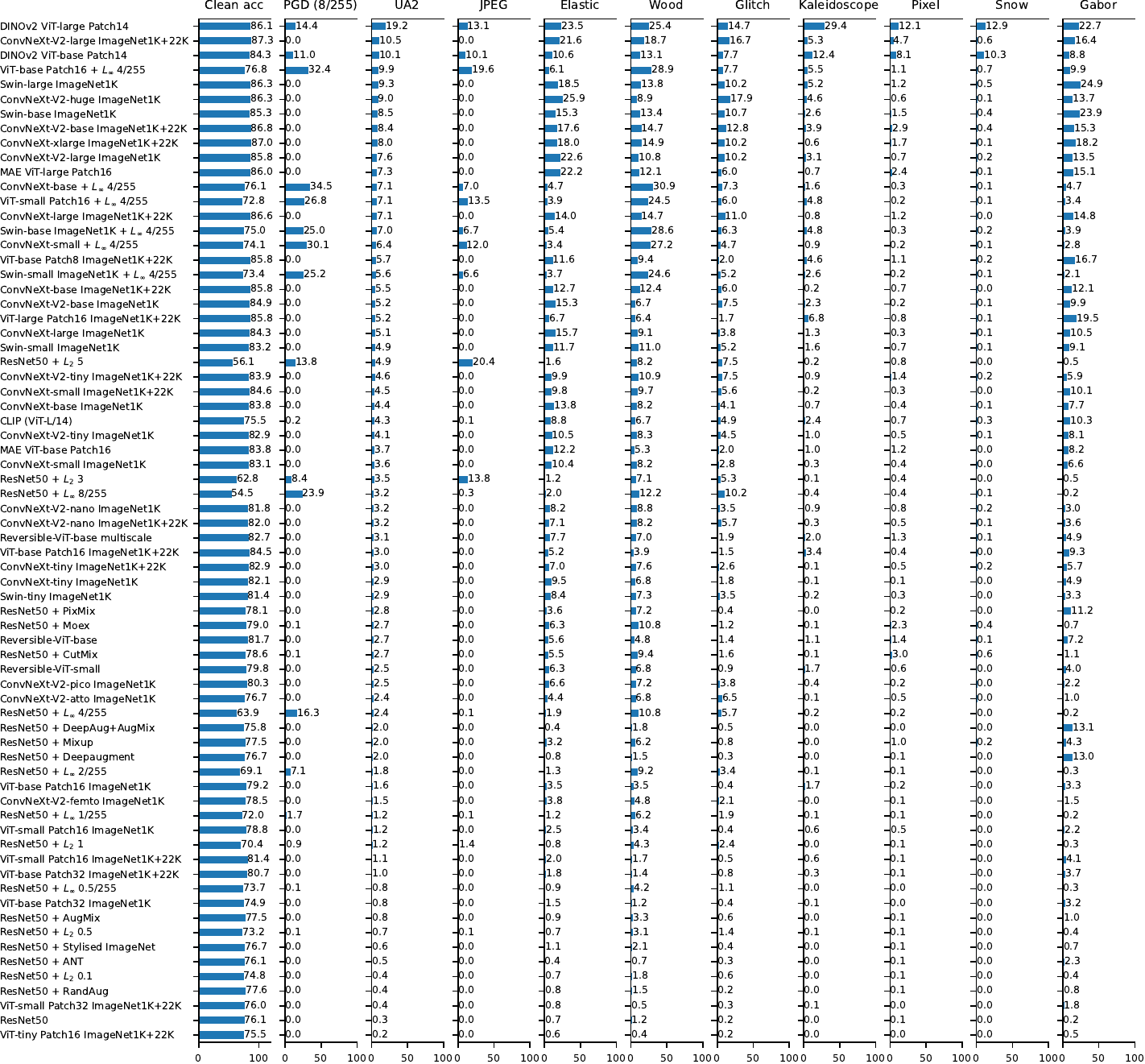}
\caption{ImageNet UA2 performance under high distortion}
\label{fig:UA2-imagenet-high}
\end{center}
\end{figure}
\clearpage

\subsection{CIFAR-10}
\begin{figure}[h!]
\begin{center}
\includegraphics[width=\textwidth]{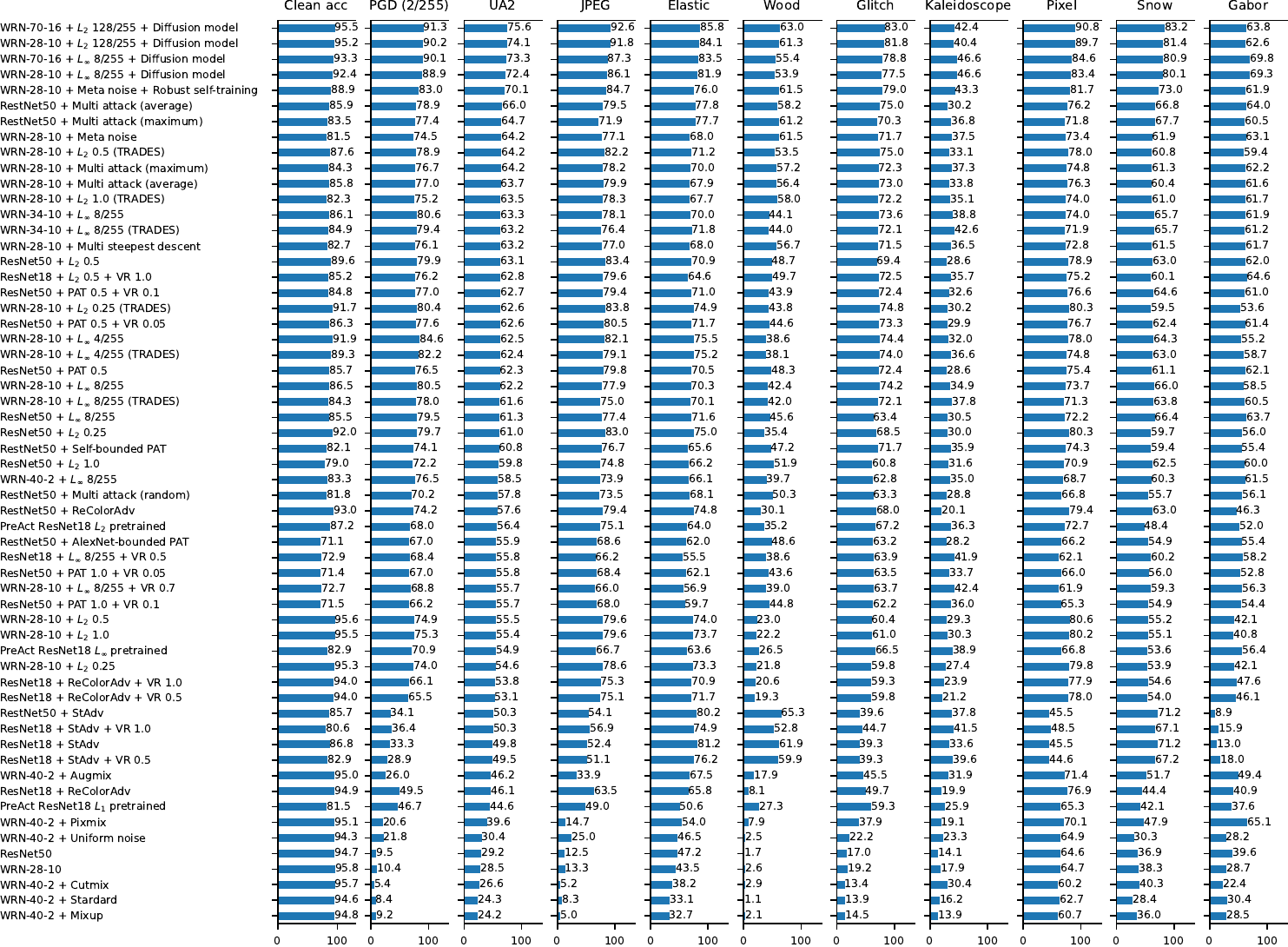}
\caption{CIFAR-10 UA2 performance under low distortion.}
\label{fig:UA2-cifar10-low}
\end{center}
\end{figure}

\begin{figure}[h!]
\begin{center}
\includegraphics[width=\textwidth]{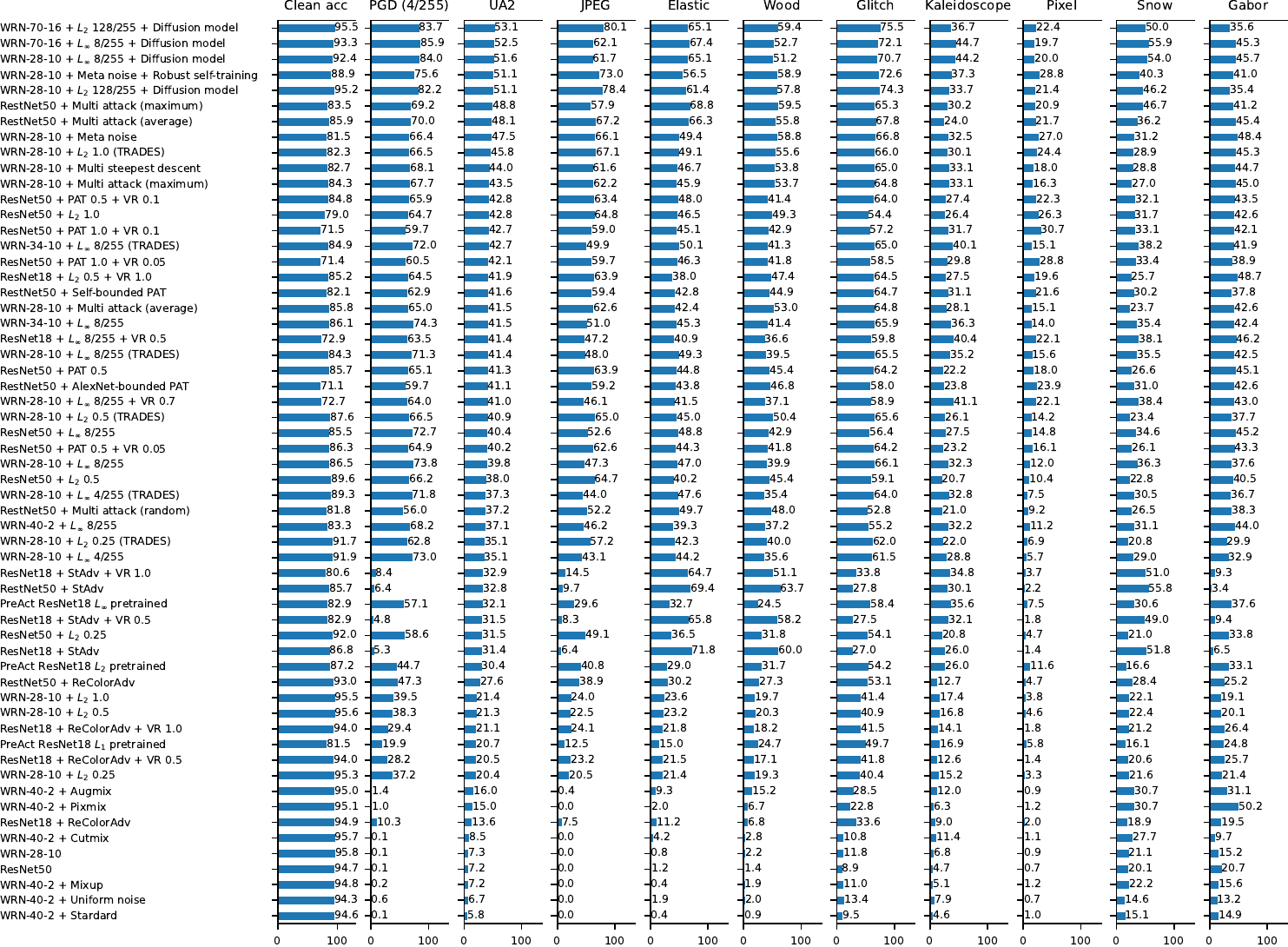}
\caption{CIFAR-10 UA2 performance under medium distortion}
\label{fig:UA2-cifar10-medium}
\end{center}
\end{figure}

\begin{figure}[h!]
\begin{center}
\includegraphics[width=\textwidth]{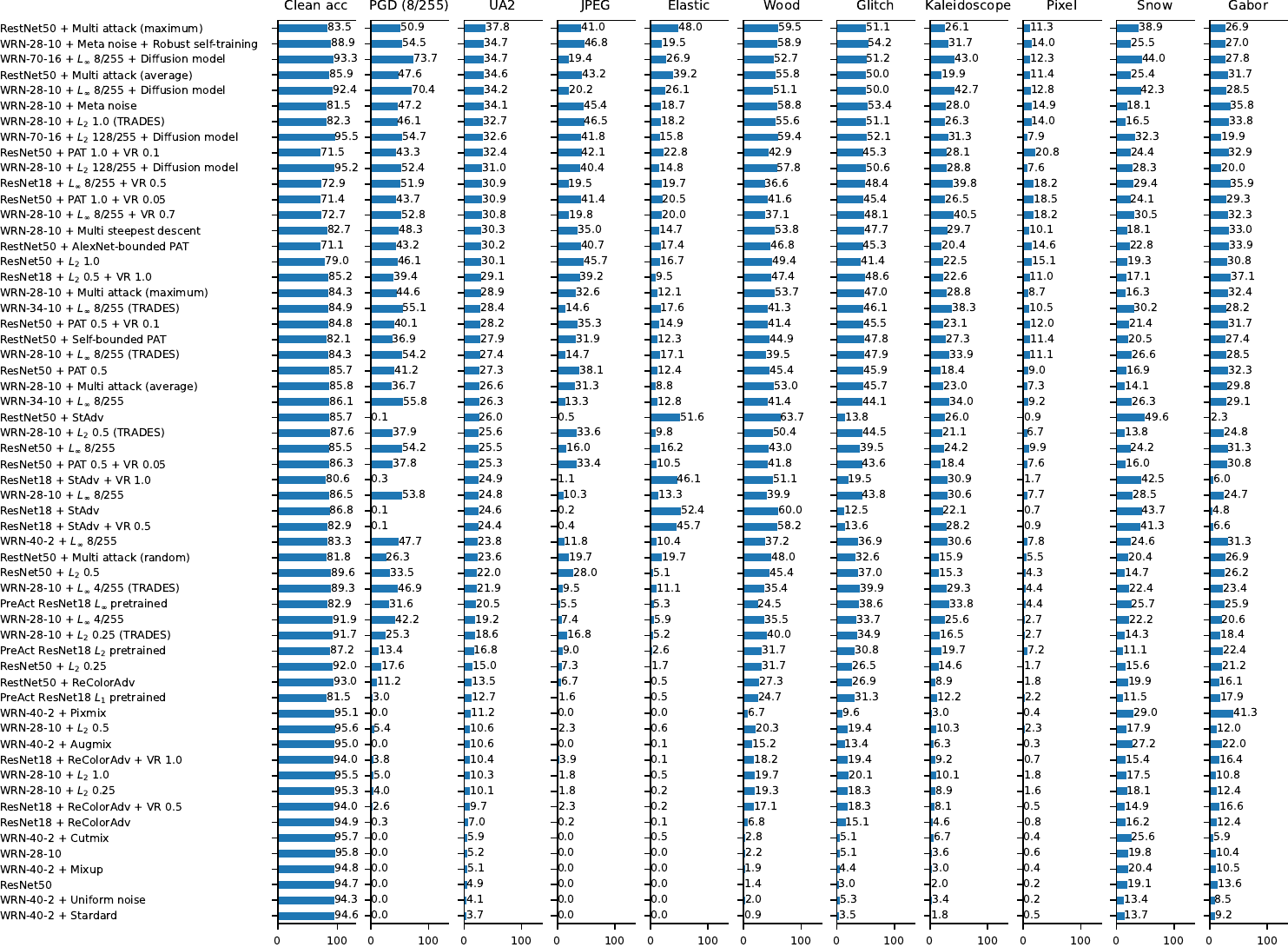}
\caption{CIFAR-10 UA2 performance under high distortion}
\label{fig:UA2-cifar10-high}
\end{center}
\end{figure}
\clearpage

\subsection{ImageNet100}
\begin{figure}[h!]
\begin{center}
\includegraphics[width=\textwidth]{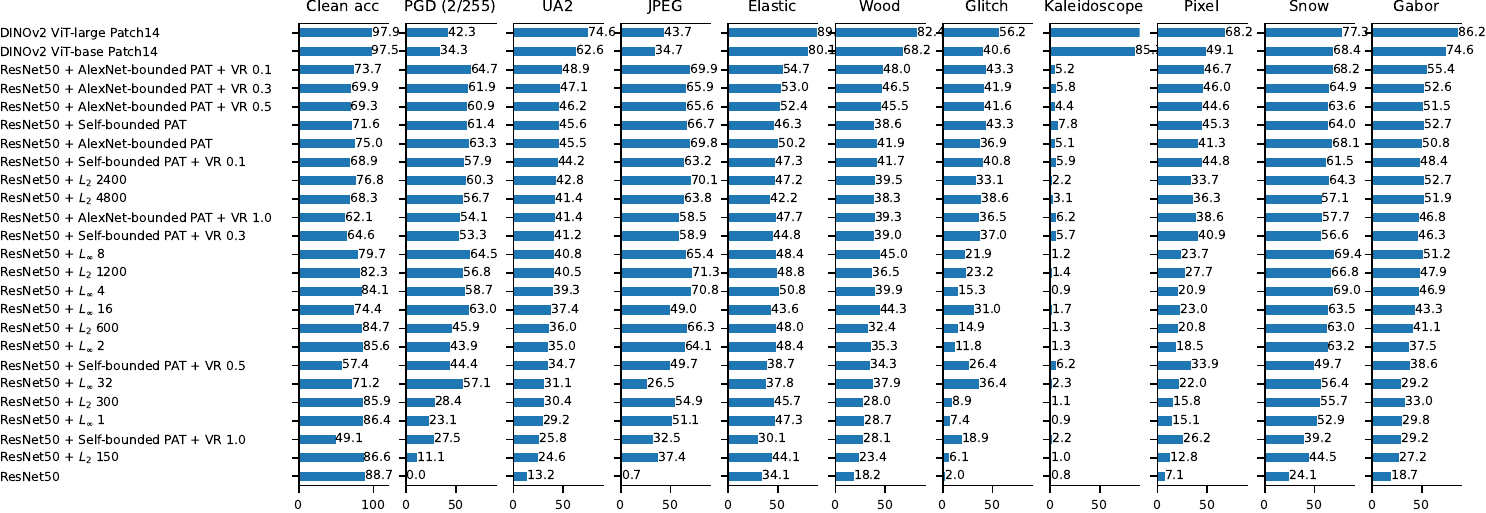}
\caption{ImageNet100 UA2 performance under low distortion.}
\label{fig:UA2-imagenet100-low}
\end{center}
\end{figure}

\begin{figure}[h!]
\begin{center}
\includegraphics[width=\textwidth]{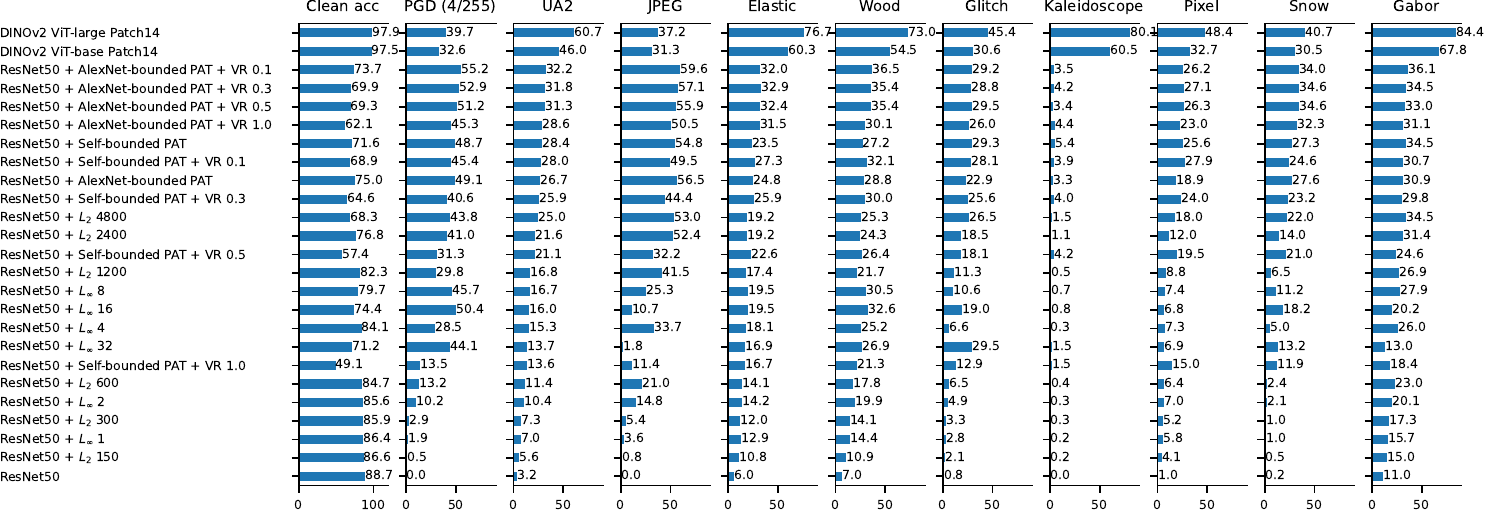}
\caption{ImageNet100 UA2 performance under medium distortion}
\label{fig:UA2-imagenet100-medium}
\end{center}
\end{figure}

\begin{figure}[h!]
\begin{center}
\includegraphics[width=\textwidth]{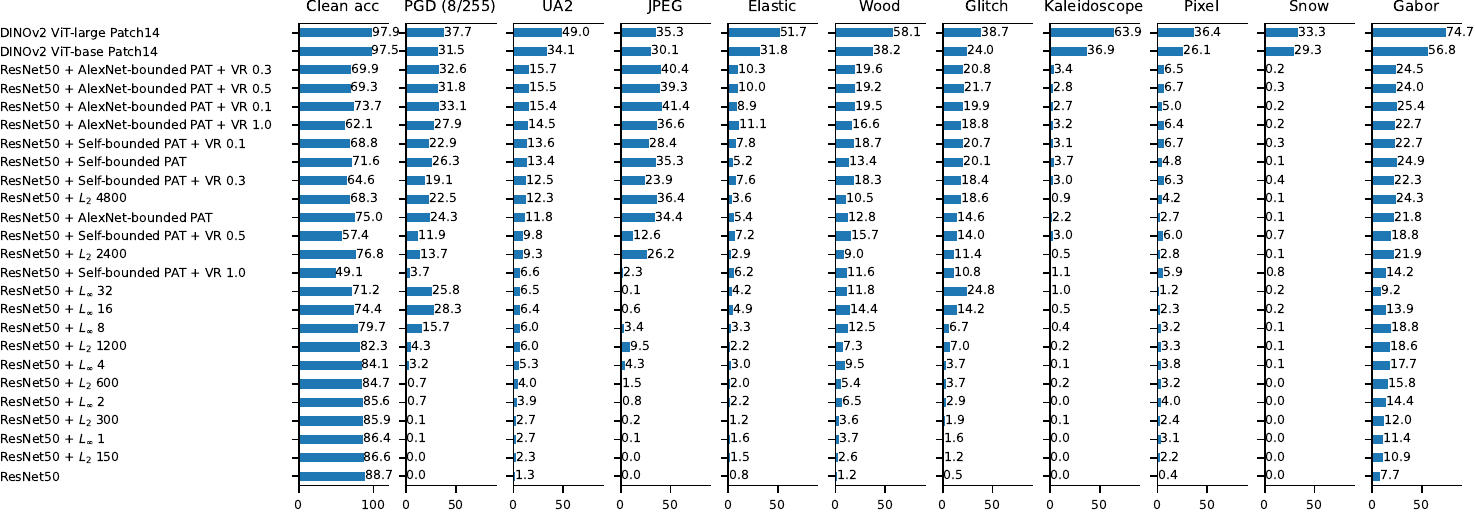}
\caption{ImageNet100 UA2 performance under high distortion}
\label{fig:UA2-imagenet100-high}
\end{center}
\end{figure}
\clearpage

\subsection{Exploring the Robustness of DINOv2}
Given the strong adversarial robustness of DINOv2 models under the PGD attack (\cref{scaling_all_attacks}), we further evaluate the DINOv2 model under AutoAttack~\cite{croce2020reliableautoattack}. \cref{tab:dinov2-robustness-imagenet} and \cref{tab:dinov2-robustness-imagenet100} show that although for the robust ResNet50 model AutoAttack performs similarly to PGD, it is able to reduce the accuracy of DINOv2 models to 0.0\% across all the distortion levels. Future work may benefite from applying the AutoAttack benchmark as a comparison point, instead of the base PGD adversary. 
\begin{table}[h!]
\centering
\label{tab:dinov2-robustness-imagenet}
\resizebox{\textwidth}{!}{%
\begin{tabular}{@{}lccc@{}}
\toprule
                         & \multicolumn{1}{l}{ResNet50 + $L_\infty$ 8/255} & \multicolumn{1}{l}{DINOv2 ViT-base Patch14} & \multicolumn{1}{l}{DINOv2 ViT-large Patch14} \\ \midrule
PGD (2/255)              & 46.8\%                                          & 12.0\%                                      & 16.7\%                                       \\
APGD-CE (2/255)          & 46.2\%                                          & 1.0\%                                       & 1.0\%                                        \\
APGD-CE + APGD-T (2/255) & 43.6\%                                          & 0.0\%                                       & 0.0\%                                        \\ \midrule
PGD (4/255)              & 38.9\%                                          & 11.4\%                                      & 15.3\%                                       \\
APGD-CE (4/255)          & 37.9\%                                          & 0.9\%                                       & 0.8\%                                        \\
APGD-CE + APGD-T (4/255) & 33.8\%                                          & 0.0\%                                       & 0.0\%                                        \\ \midrule
PGD (8/255)              & 23.9\%                                          & 11.0\%                                      & 14.4\%                                       \\
APGD-CE (8/255)          & 22.6\%                                          & 0.6\%                                       & 0.7\%                                        \\
APGD-CE + APGD-T (8/255) & 18.4\%                                          & 0.0\%                                       & 0.0\%                                        \\ \bottomrule
\end{tabular}%
}
\caption{Attacked accuracies of models on ImageNet}
\end{table}
\begin{table}[h!]
\centering

\label{tab:dinov2-robustness-imagenet100}
\resizebox{\textwidth}{!}{%
\begin{tabular}{@{}lccc@{}}
\toprule
                         & \multicolumn{1}{l}{ResNet50 + $L_\infty$ 8/255} & \multicolumn{1}{l}{DINOv2 ViT-base Patch14} & \multicolumn{1}{l}{DINOv2 ViT-large Patch14} \\ \midrule
PGD (2/255)              & 64.5\%                                          & 34.3\%                                      & 42.3\%                                       \\
APGD-CE (2/255)          & 64.4\%                                          & 17.6\%                                      & 20.0\%                                       \\
APGD-CE + APGD-T (2/255) & 64.1\%                                          & 0.0\%                                       & 0.0\%                                        \\ \midrule
PGD (4/255)              & 45.7\%                                          & 32.6\%                                      & 39.7\%                                       \\
APGD-CE (4/255)          & 45.2\%                                          & 16.4\%                                      & 17.3\%                                       \\
APGD-CE + APGD-T (4/255) & 44.6\%                                          & 0.0\%                                       & 0.0\%                                        \\ \midrule
PGD (8/255)              & 15.7\%                                          & 31.5\%                                      & 37.7\%                                       \\
APGD-CE (8/255)          & 14.7\%                                          & 15.5\%                                      & 14.5\%                                       \\
APGD-CE + APGD-T (8/255) & 13.6\%                                          & 0.0\%                                       & 0.0\%                                        \\ \bottomrule
\end{tabular}%
}
\caption{Attacked accuracies of models on ImageNet100}
\end{table}

\subsection{Performance Variance}
As described in \cref{generation-strategy-attacks}, we perform adversarial attacks by optimizing latent variables which are randomly initialized in our current implementation, so the adversarial attack's performance can be affected by the random seed for the initialization. To study the effect of random initializations, we compute the UA2 performances of three samples of two ImageNet models, ResNet50 and ResNet50 + $L_2$ 5. We observe the standard deviations of UA2 of these two models across 5 different seeds to be respectively 0.1\% and 0.04\% concluding that the variation in performance across the ImageNet dataset is minor.

\section{Images of All Attacks Across Distortion Levels}
\label{app:attack-images}
We provide images of all 19 attacks within the benchmark, across the three distortion levels.
\begin{figure}
    \centering
Gabor \includegraphics[width=\linewidth]{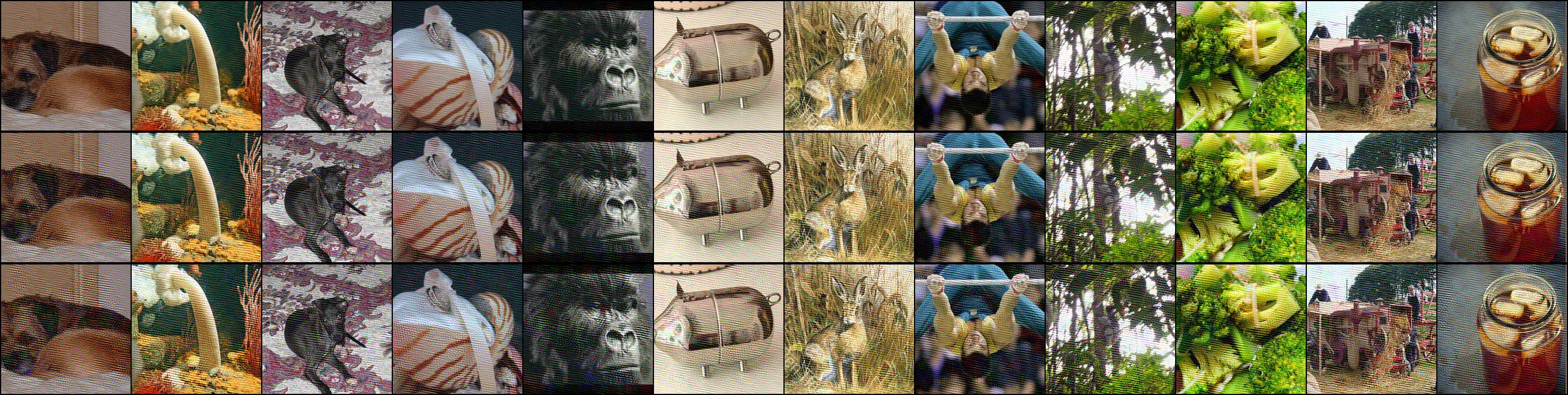}
Kaleidoscope \includegraphics[width=\linewidth]{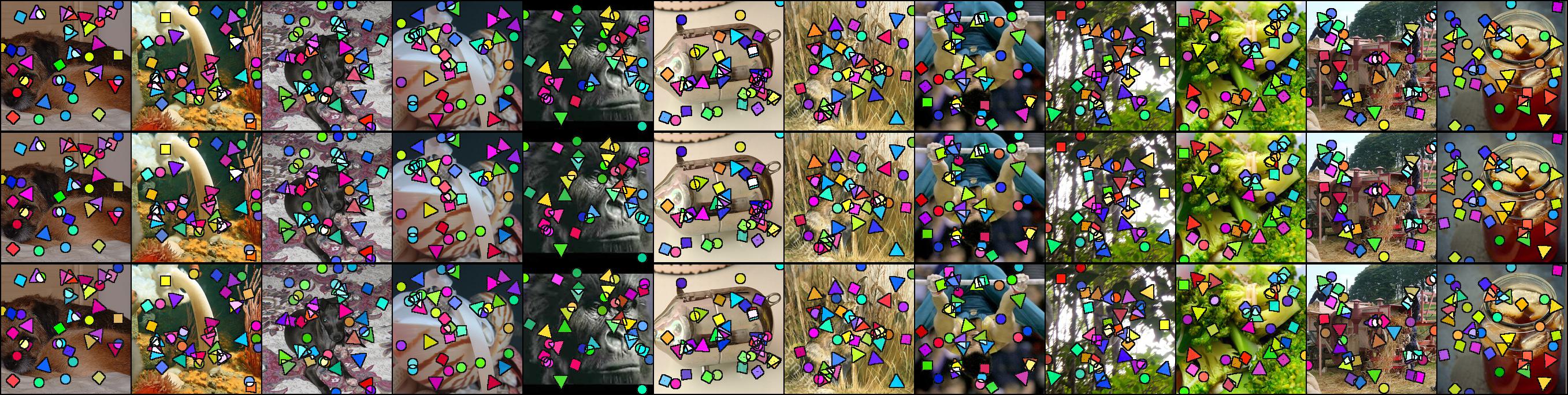}
JPEG \includegraphics[width=\linewidth]{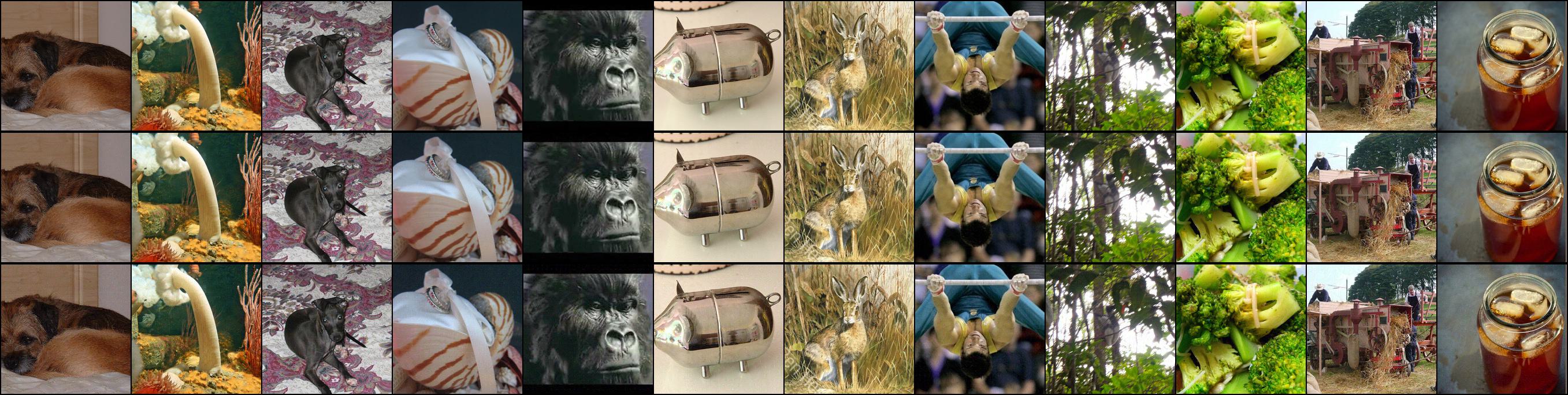}
Mix \includegraphics[width=\linewidth]{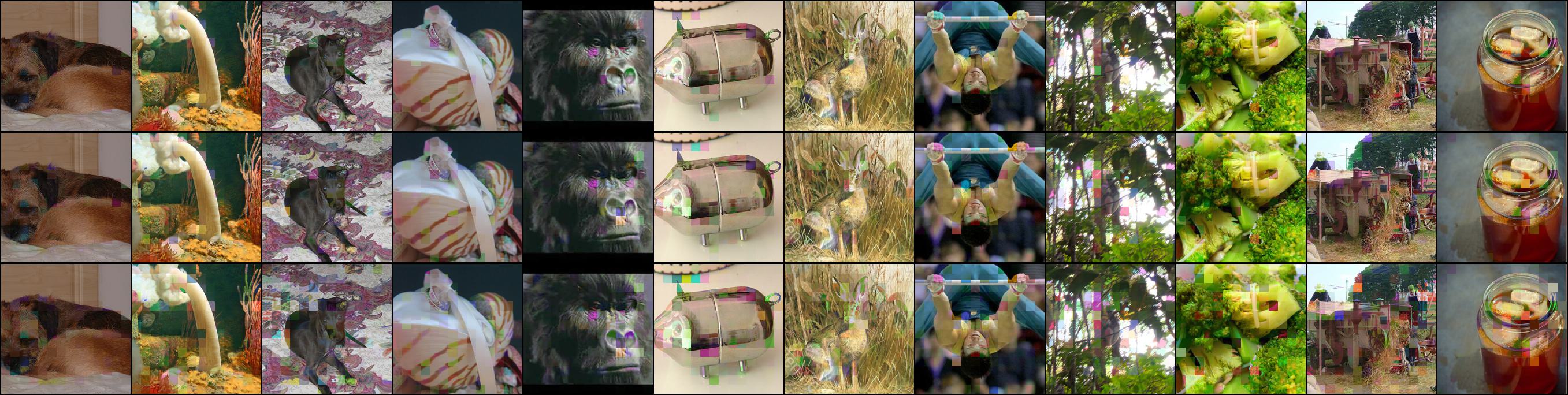}
Snow \includegraphics[width=\linewidth]{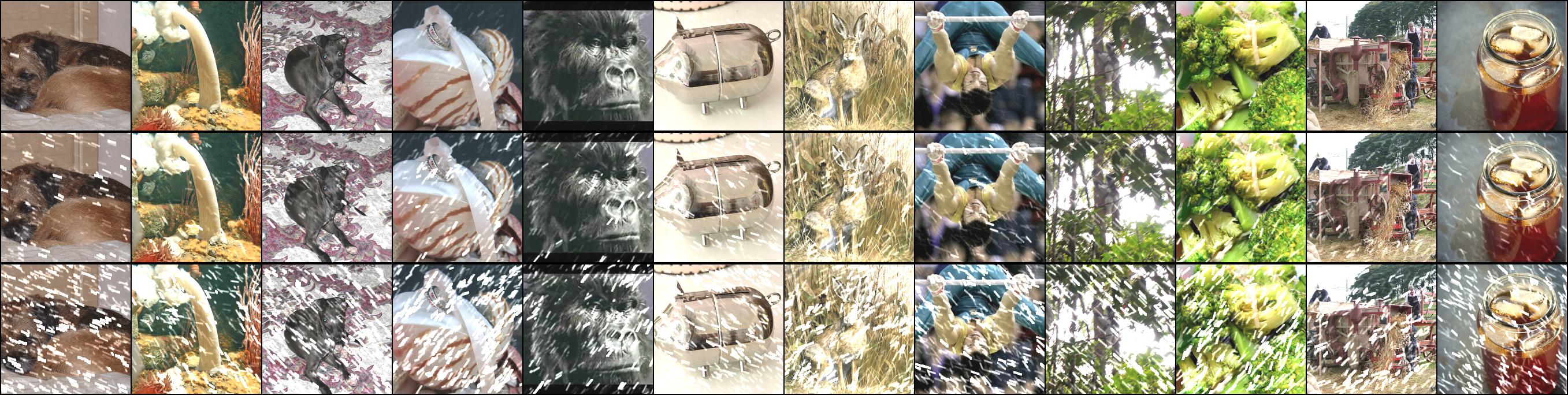}
    \caption{Attacked samples of low distortion (1st row), medium distortion (2nd row), and high distortion (last row) on a standard ResNet50 model}

\end{figure}

\begin{figure}
    \centering
Glitch \includegraphics[width=\linewidth]{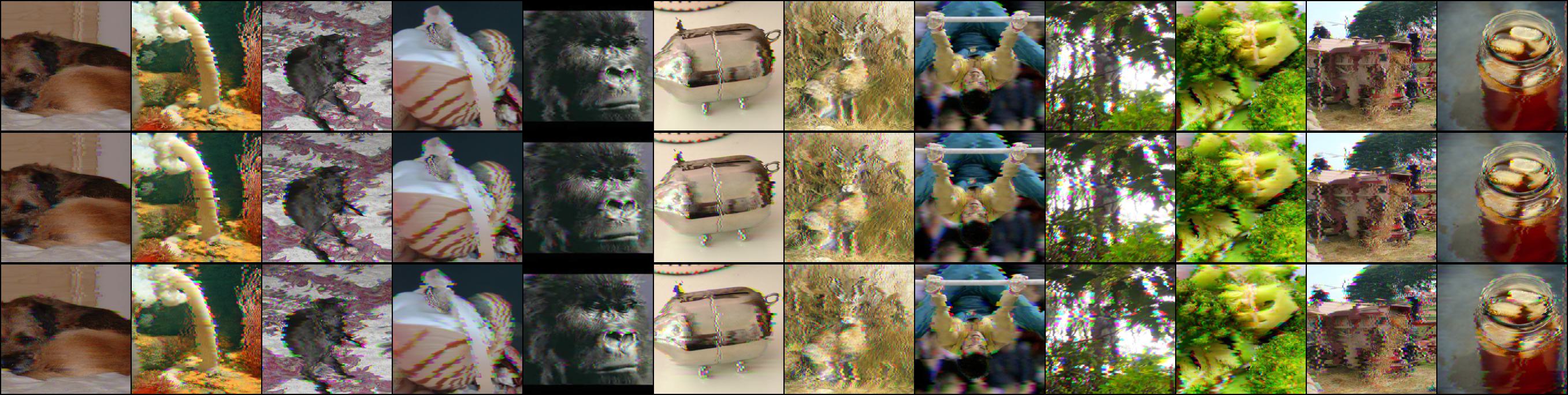}
Wood \includegraphics[width=\linewidth]{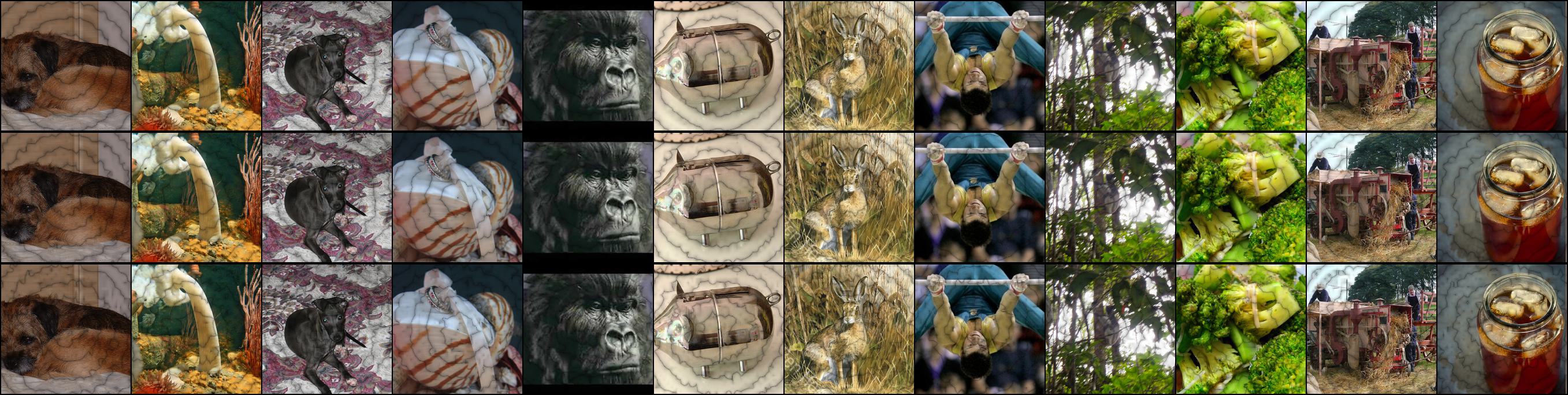}
Elastic \includegraphics[width=\linewidth]{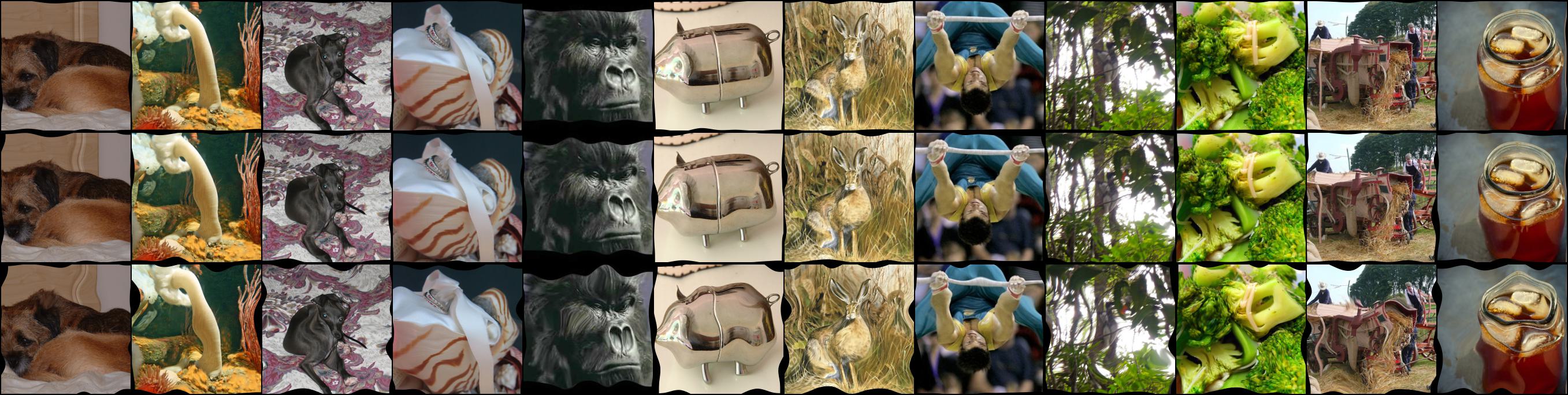}
Edge \includegraphics[width=\linewidth]{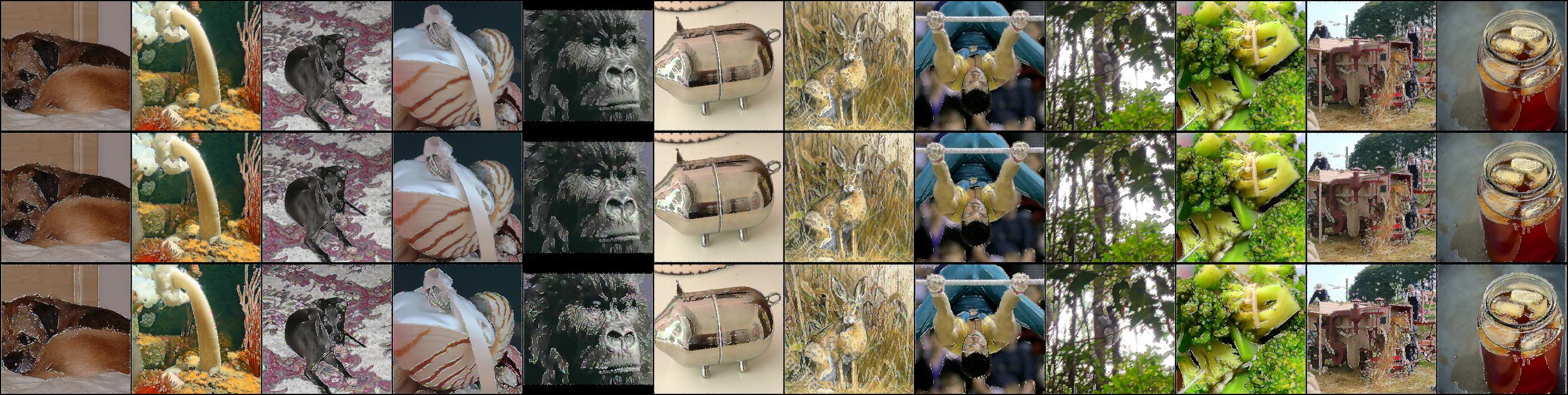}
FBM \includegraphics[width=\linewidth]{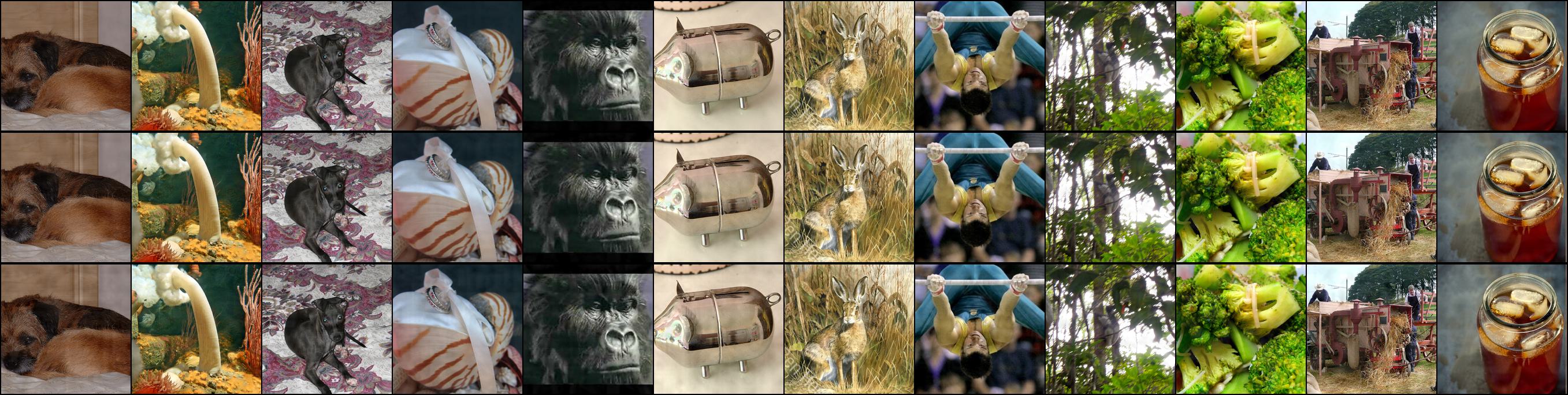}
    \caption{Attacked samples of low distortion (1st row), medium distortion (2nd row), and high distortion (last row) on a standard ResNet50 model}

\end{figure}
\begin{figure}
    \centering
Fog \includegraphics[width=\linewidth]{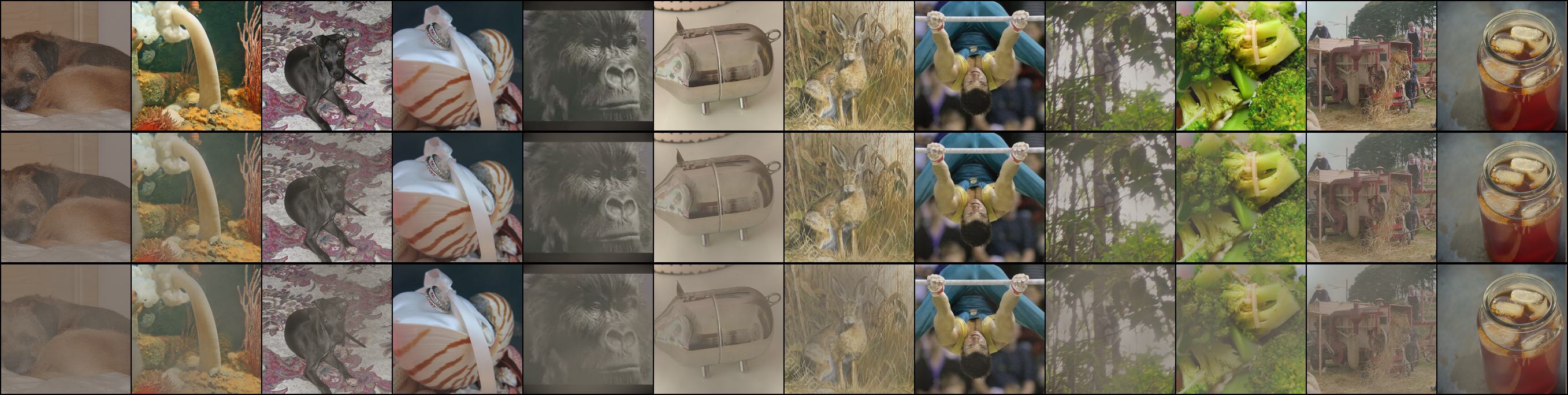}
HSV \includegraphics[width=\linewidth]{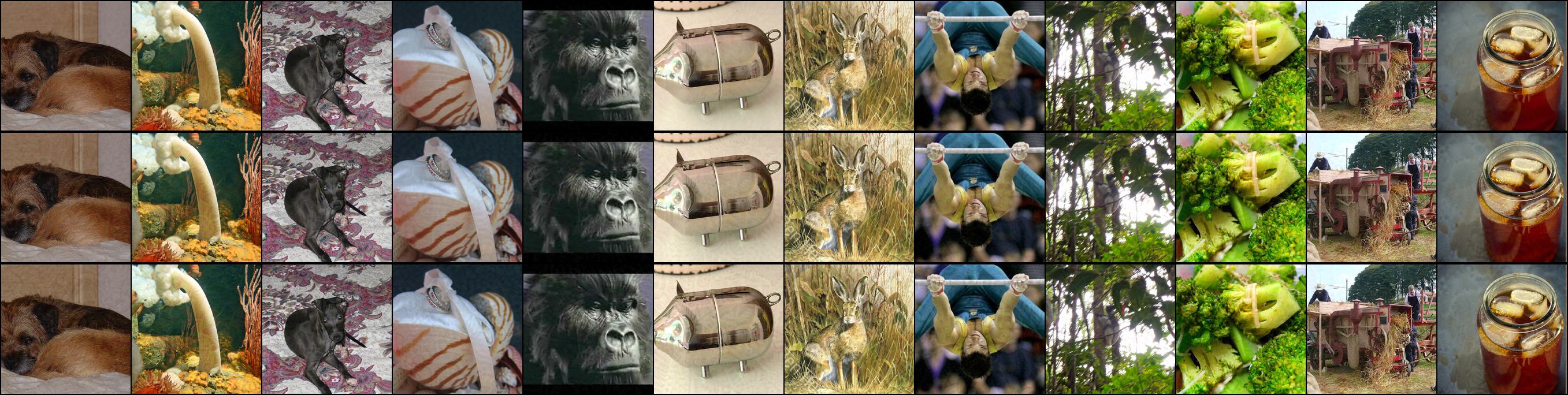}
Klotski \includegraphics[width=\linewidth]{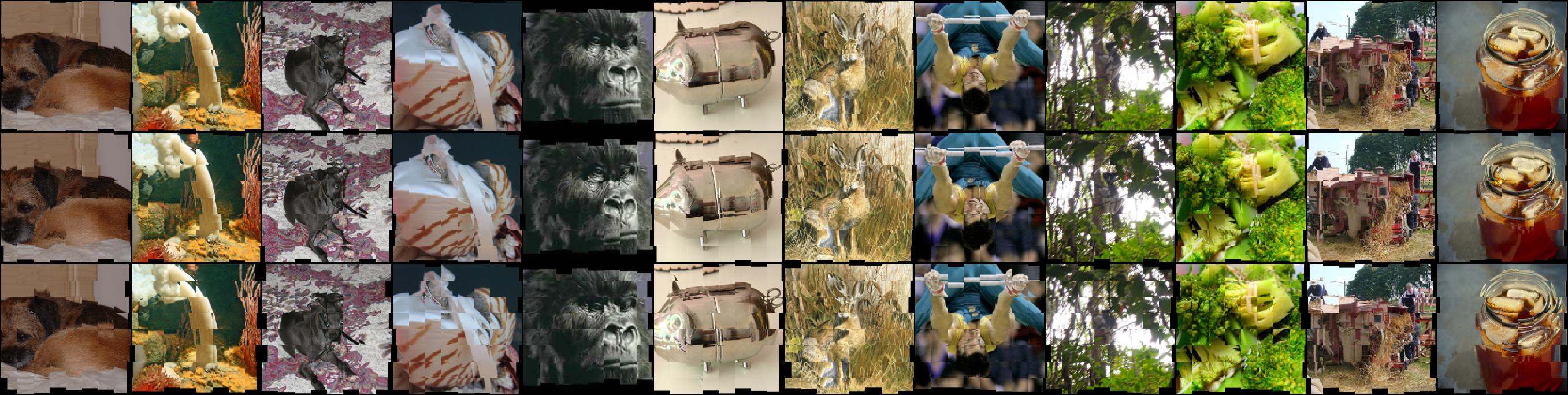}
Mix \includegraphics[width=\linewidth]{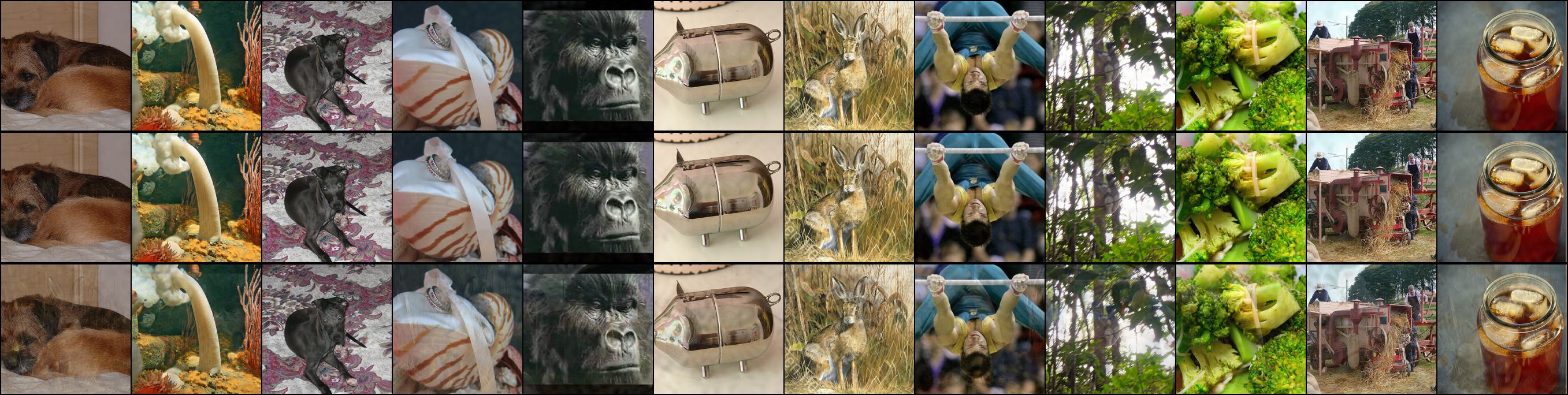}
Polkadot \includegraphics[width=\linewidth]{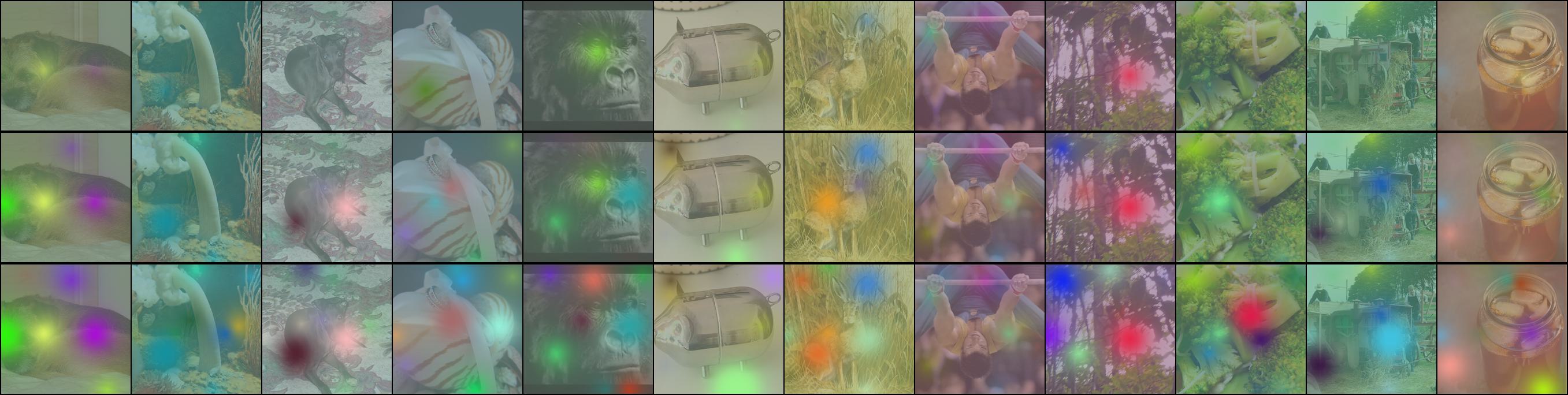}
    \caption{Attacked samples of low distortion (1st row), medium distortion (2nd row), and high distortion (last row) on a standard ResNet50 model}

\end{figure}
\begin{figure}
    \centering
Prison \includegraphics[width=\linewidth]{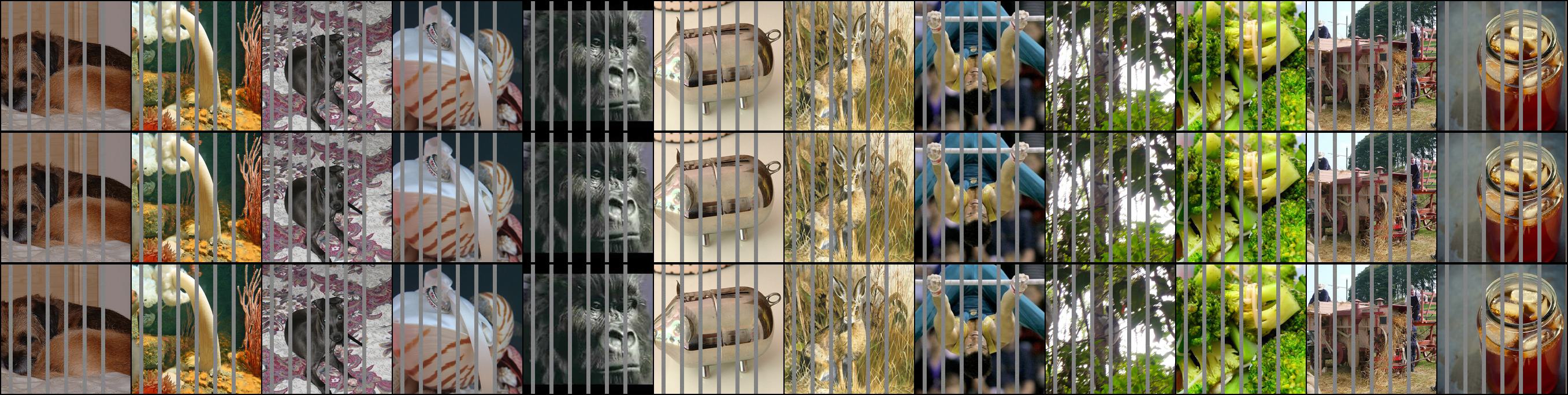}
Blur \includegraphics[width=\linewidth]{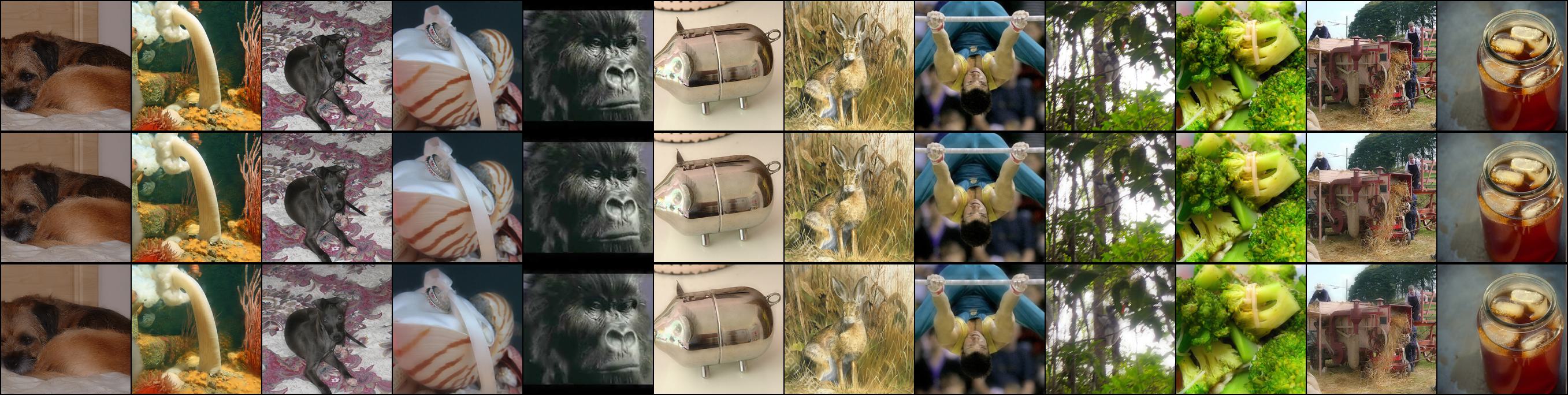}
Texture \includegraphics[width=\linewidth]{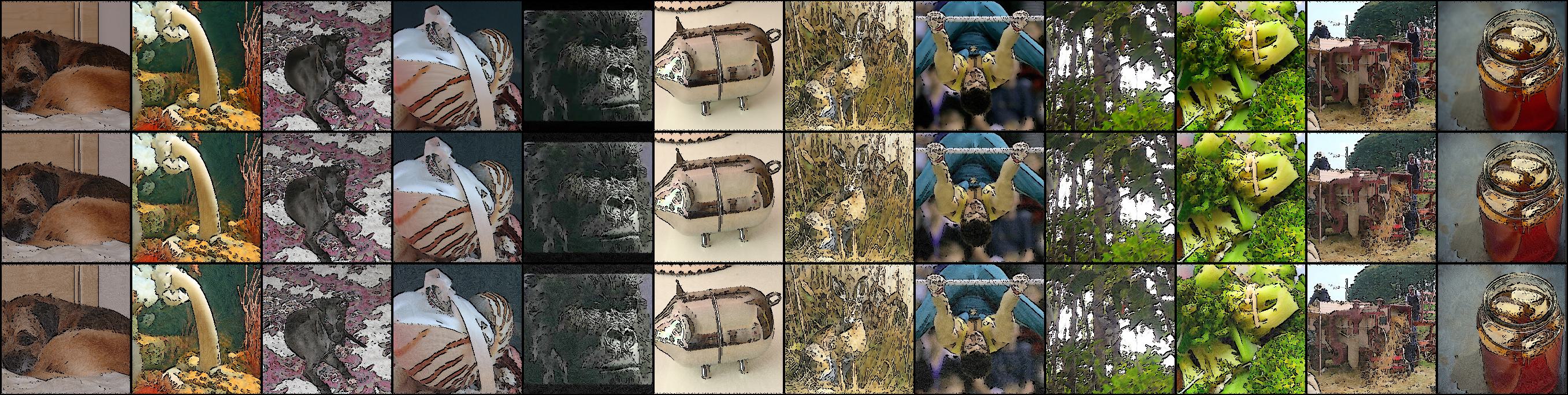}
Whirlpool \includegraphics[width=\linewidth]{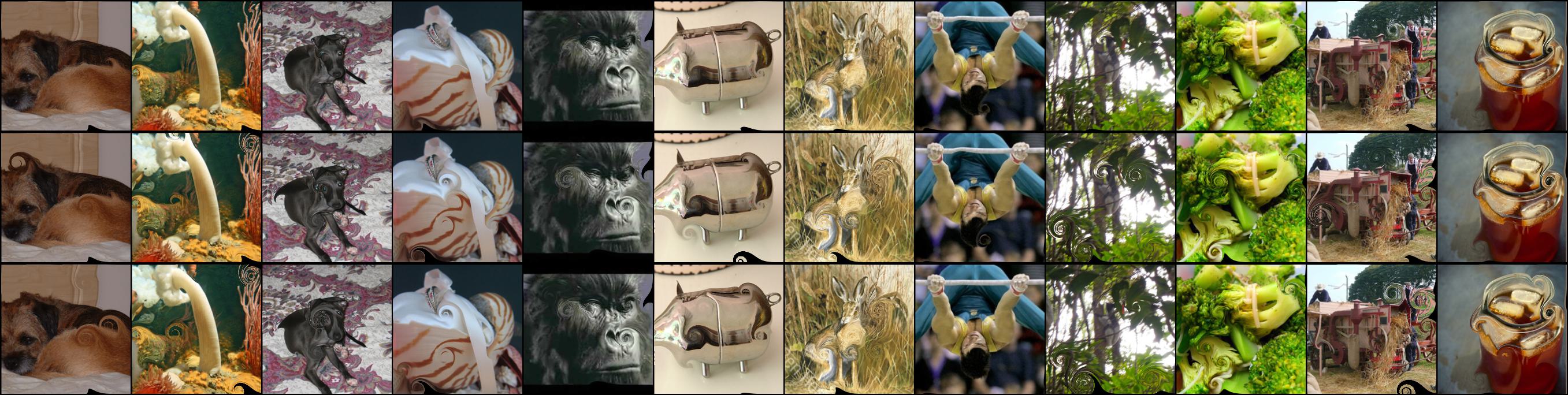}
    \caption{Attacked samples of low distortion (1st row), medium distortion (2nd row), and high distortion (last row) on a standard ResNet50 model}

\end{figure}

\FloatBarrier
\section{Scaling Behaviour of Our Attacks}
To see how our attacks perform across model scale, we make use of the ConvNeXt-V2 model suite \citep{woo2023convnext} to test the performance of our attacks as we scale model size. We find that capacity improves performance across the board, but find diminishing returns to simply scaling up the architectures, pointing towards techniques described in \cref{app:non-lp-training}.

\label{app:scaling-convnext-results}
    \begin{figure}[ht]
        \centering
        \includegraphics[width=.8\linewidth]{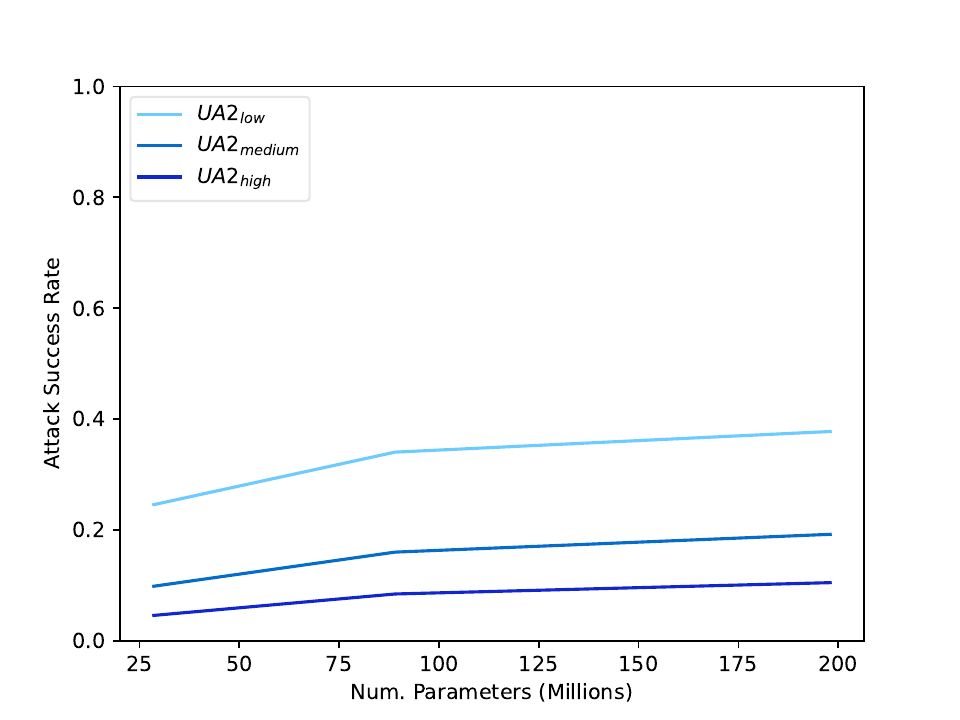}
        \caption{\textbf{Unforeseen Robustness across model scale.} We measure $\UAA$ across model scale by evaluating the performance of ConvNeXt-V2 \citep{woo2023convnext} models on $\task$, finding that scale improves performance, although the benchmark still provides a challenge to the largest models.}
    \end{figure}

\begin{figure}[!ht]
    \centering
    
    \begin{subfigure}{.5\textwidth}
        \centering
        \includegraphics[width=.8\linewidth]{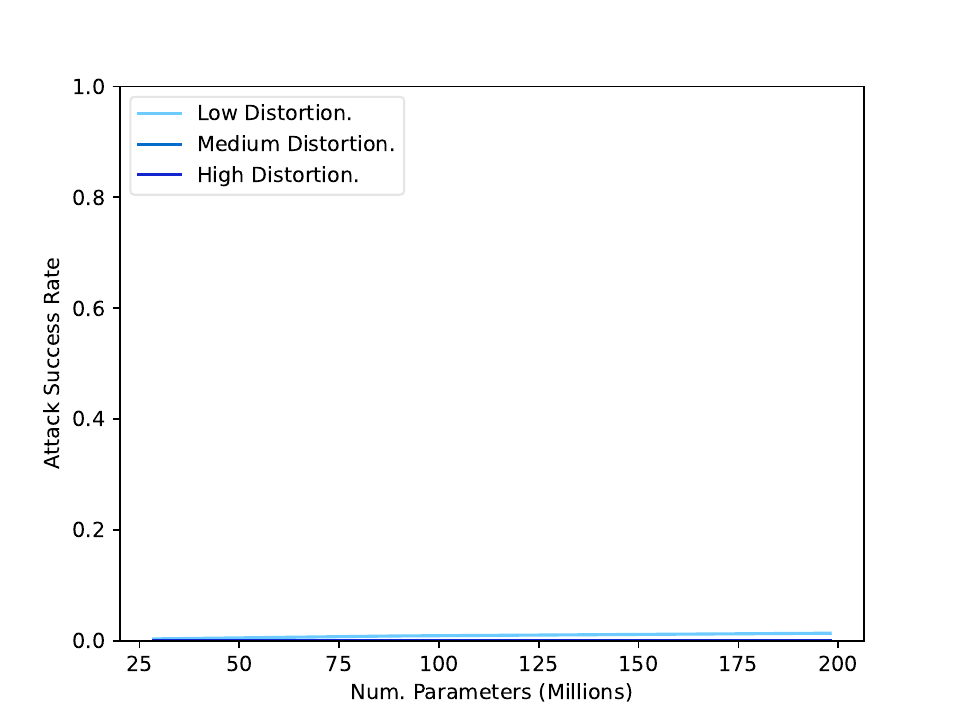}
        \caption{JPEG}
    \end{subfigure}%
    \begin{subfigure}{.5\textwidth}
        \centering
        \includegraphics[width=.8\linewidth]{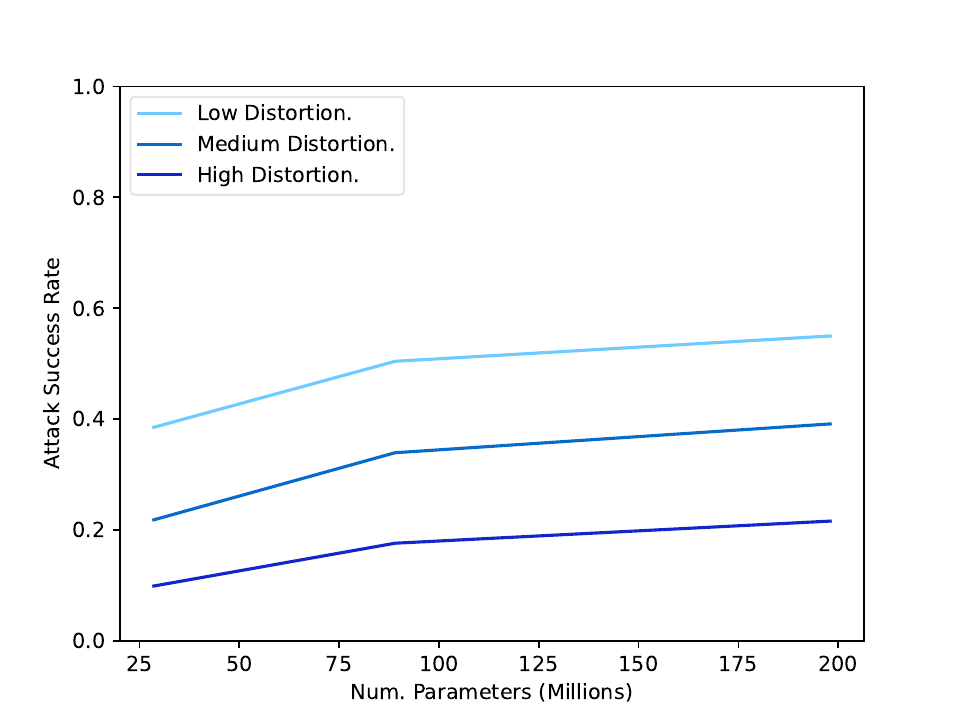}
        \caption{Elastic}
    \end{subfigure}

    \begin{subfigure}{.5\textwidth}
        \centering
        \includegraphics[width=.8\linewidth]{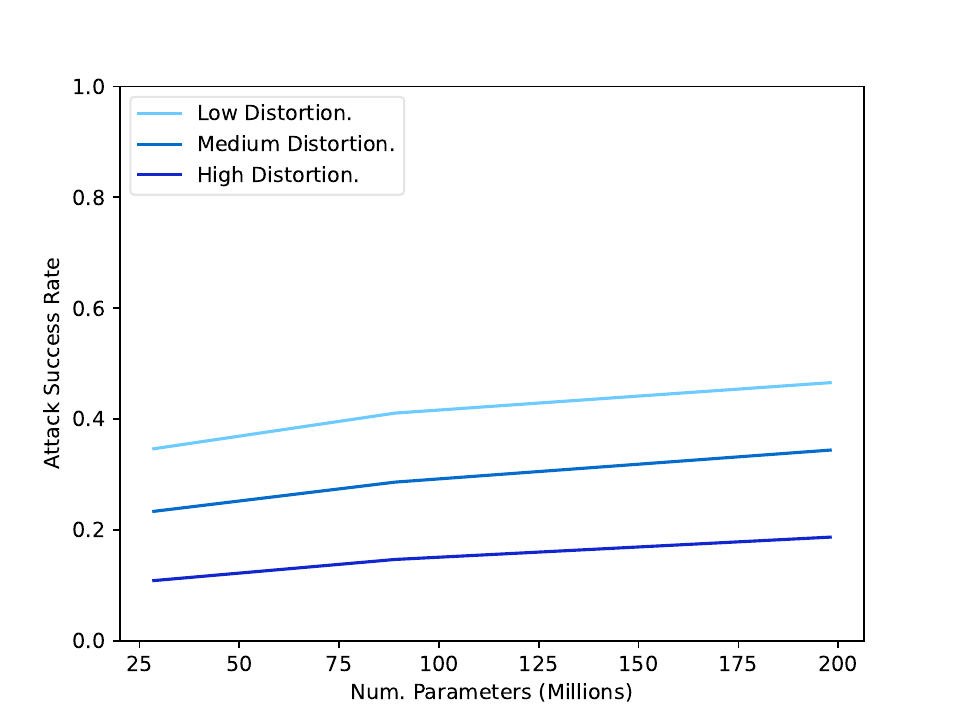}
        \caption{Wood}
    \end{subfigure}%
    \begin{subfigure}{.5\textwidth}
        \centering
        \includegraphics[width=.8\linewidth]{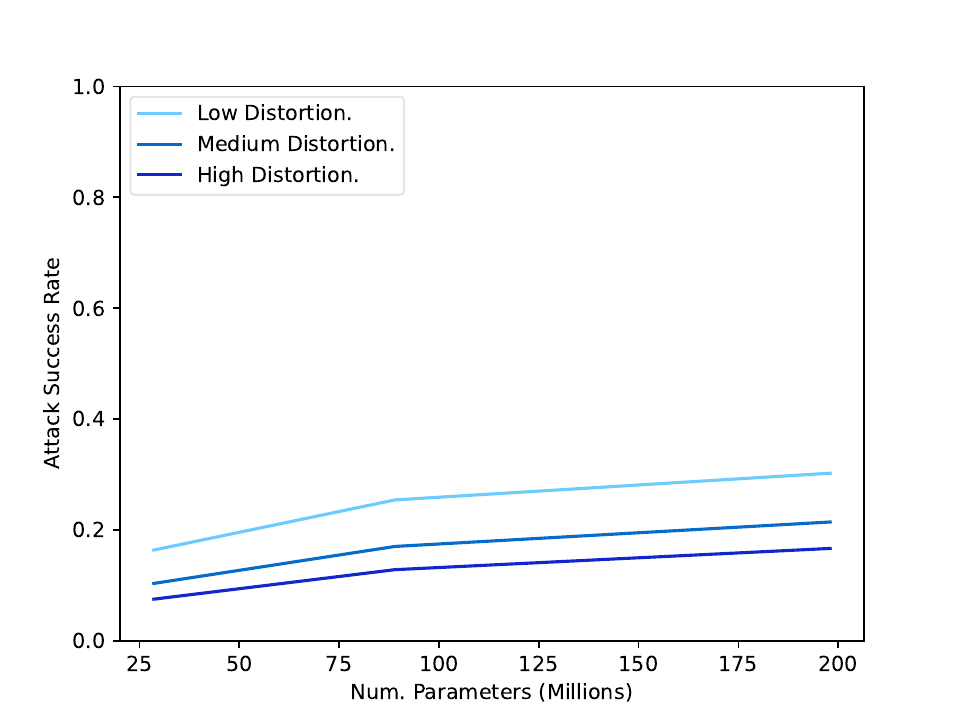}
        \caption{Glitch}
    \end{subfigure}

    \begin{subfigure}{.5\textwidth}
        \centering
        \includegraphics[width=.8\linewidth]{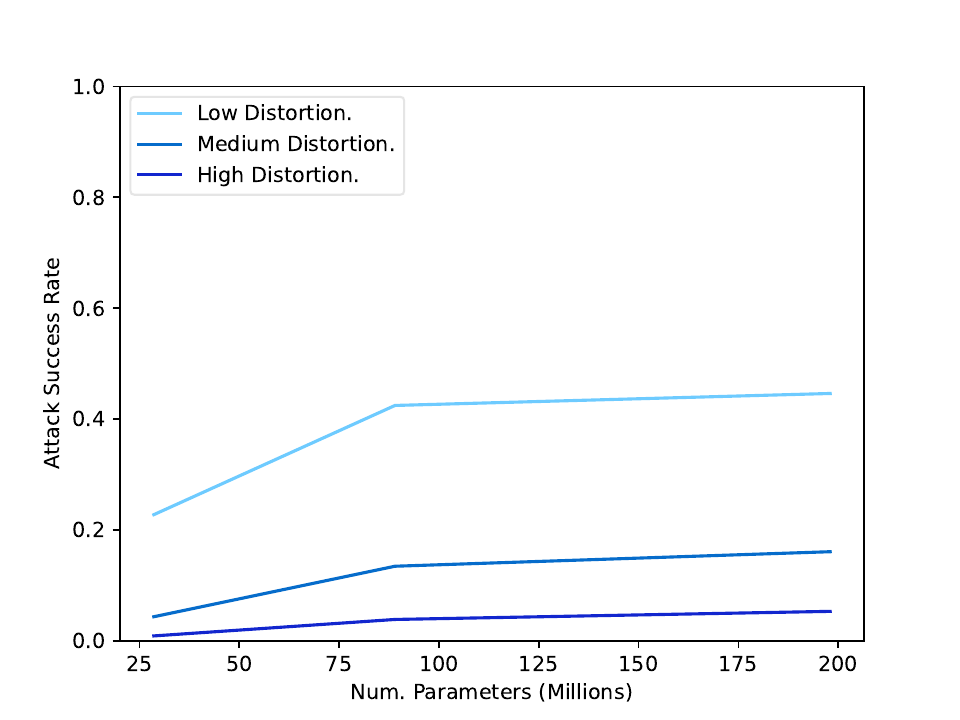}
        \caption{Kaleidoscope}
    \end{subfigure}%
    \begin{subfigure}{.5\textwidth}
        \centering
        \includegraphics[width=.8\linewidth]{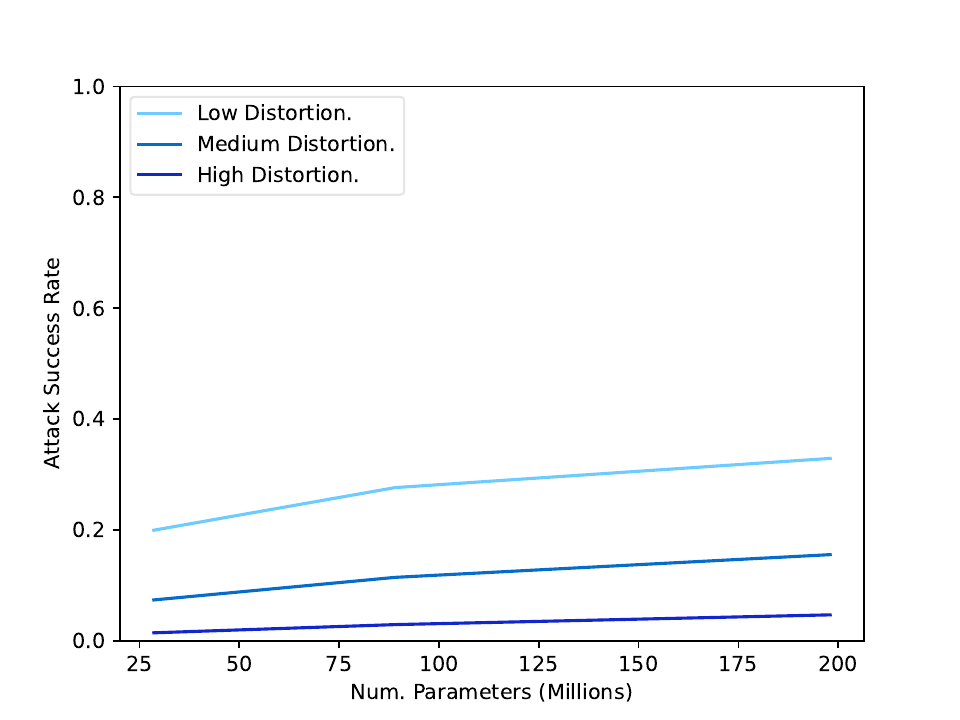}
        \caption{Pixel}
    \end{subfigure}

    \begin{subfigure}{.5\textwidth}
        \centering
        \includegraphics[width=.8\linewidth]{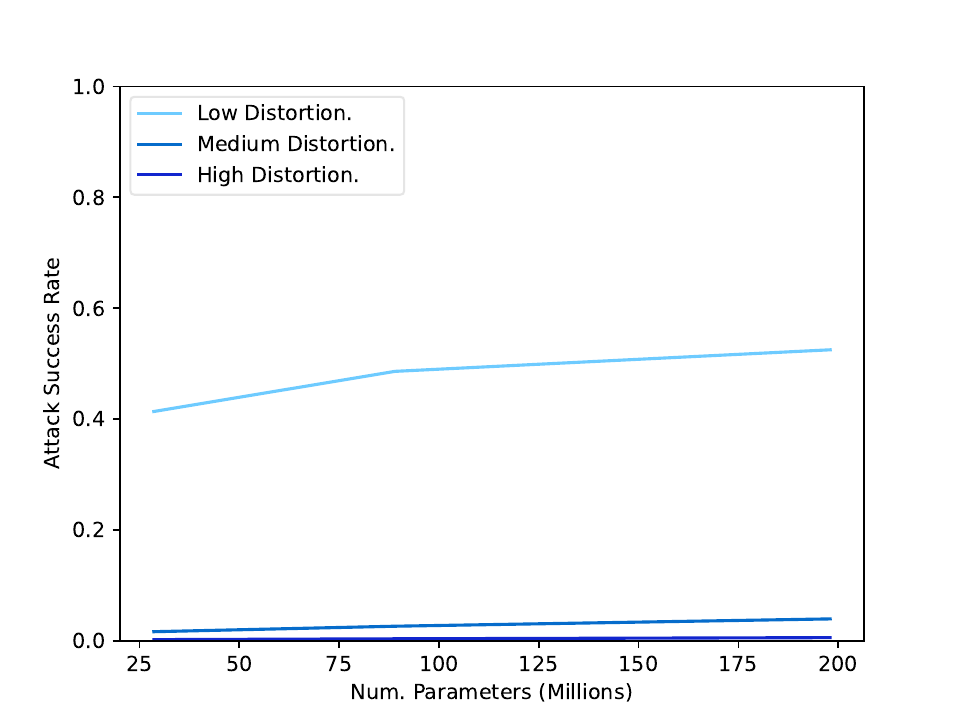}
        \caption{Snow}
    \end{subfigure}%
    \begin{subfigure}{.5\textwidth}
        \centering
        \includegraphics[width=.8\linewidth]{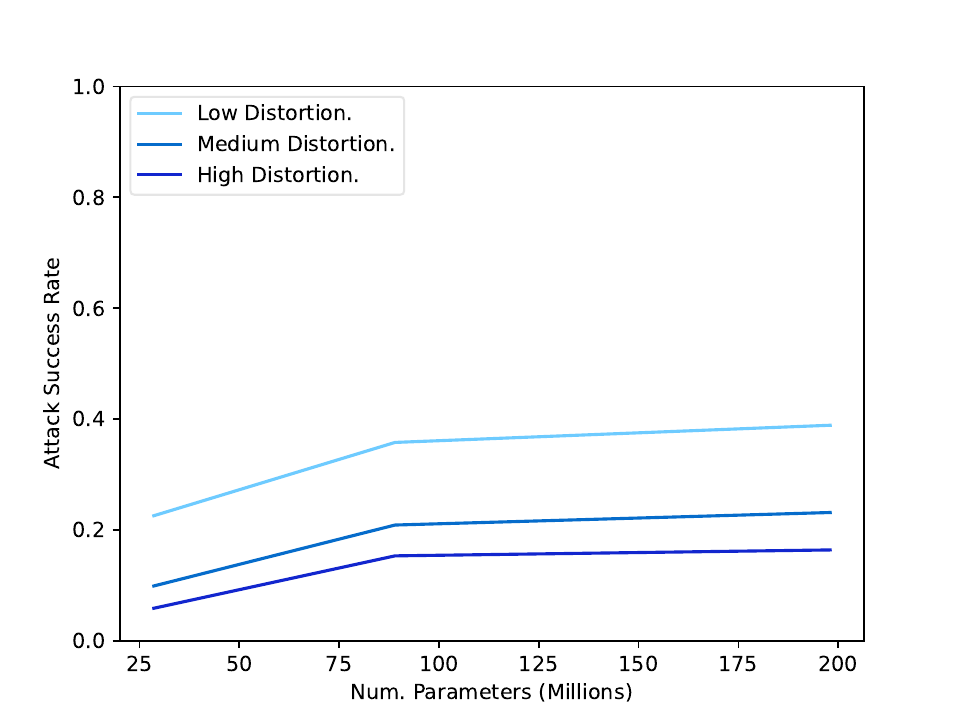}
        \caption{Gabor}
    \end{subfigure}

    \caption{\textbf{Behaviour of core attacks across model scale.} We see the performance of the eight core attacks across the ConvNeXt-V2 model suite, with performance on attacks improving with model scale.\vspace{5cm}}
    \label{fig:core_attacks}
\end{figure}

\section{ImageNet-C and unforeseen robustness}
\label{app:imagenet-c-unforeseen}
\begin{table}[!htp]

\label{tab:common-corruptions-different}
\centering
\begin{tabular}{lccc}\toprule
Model & $\UAA$ (non-optimized) $\uparrow$ & mCE  $\downarrow$ & UA2  $\uparrow$  \\\midrule
Resnet 50 & 55.2 & 76.7 & 1.6 \\
Resnet50 + AugMix & 59.1 & 65.7 & 3.5 \\
Resnet50 + DeepAug & 60.2 & \textbf{61.1} & 3.0 \\
Resnet50 + Mixup & 59.9 & 69.2 & 4.8 \\\midrule
Resnet50 + $L_2$, ($\varepsilon = 5$) & 43.2 & 89.0 & 13.9 \\
Resnet50 + $L_{\infty}$, ($\varepsilon = 8/255$) & 40.6 & 85.1 & 10 \\ \bottomrule \\
\end{tabular}
\caption{\textbf{Common corruptions and UA2} We compare performance on the ImageNet-C benchmark (mCE) to performance against both non-optimized and optimized versions of our attacks. We find that performance on the average-case robustness of ImageNet-C is correlated with performance on optimised attacks, while applying optimised versions favours the adversarially trained models.}
\end{table}

\FloatBarrier
\section{Benchmarking Non- \texorpdfstring{$L_p$} \space\space Adversarial Training Strategies}
\label{app:non-lp-training}

We wish to compare training strategies which have been specifically developed for robustness against both a variety of and unforeseen adversaries. To this end, we use Meta Noise Generation \citep{madaan2021learninggeneratenoise} as a strong multi-attack robustness baseline, finding that on $\cifartask$ this leads to large increases in robustness (\cref{tab:meta-noise-generation}). We also evaluate Perceptual Adversarial Training  \citep{perceptualdistance} and Variational Regularization \citep{dai2022formulating}, two techniques specifically designed to achieve unforeseen robustness. We also evaluate combining PixMix and $L_p$ adversarial training. All of these baselines beat $L_p$ training.

 \begin{table}[!ht]  
    \label{tab:multi-attack-robustness-table}
 \centering
    \label{tab:meta-noise-generation}
    \begin{tabular}{ccc}\toprule
     Training  & Clean Acc. &UA2 \\
     \midrule
     Standard & \textbf{95.8} &7.4\\
     $L_\infty, \varepsilon$ = $8/255$ &86.5 &39.8 \\
     $L_2, \varepsilon = 2$  & 95.5 & 21.4\\
     MNG &  88.9 & \textbf{51.1}\\
    \bottomrule\\
    \end{tabular}
         \caption{\textbf{Comparing alternative training strategies to $L_p
    $ baselines} We demonstrate that models trained using Meta Noise Generation (MNG) \citep{madaan2021learninggeneratenoise} improve over $L_p$ training baselines on $\cifartask$.}
\end{table}

\textbf{Meta Noise Generation (MNG) out-performs $L_p$ baselines.} We find that MNG, a technique original developed for multi-attack robustness shows a 11.3\% increase in $\UAA$ on $\cifartask$, and PAT shows a $3.5\%$ increase in $\UAA$.
\begin{table}[!ht]

\label{tab:perceptual-adversarial-training}
\centering
    \label{tab:pat-results}
    \begin{tabular}{ccccc}\toprule
    Training  & Clean Acc. &UA2 \\
    \midrule
    Standard &\textbf{88.7} &3.2\\ 
    $L_\infty$, $\varepsilon =  8/255$ &79.7 &17.5\\
    $L_2$, $\varepsilon = 4800/255$ &71.6 &25.0   \\
    \midrule
    PAT & 75.0  & 26.2  \\
    PAT-VR & 69.4 & \textbf{29.5}   \\
    \bottomrule\\
    \end{tabular}
\caption{ \textbf{Specialised Unforseen robustness training strategies.} We see that $\task$ PAT \citep{perceptualdistance} and PAT-VR \citep{dai2022formulating}trained ResNet50s improve over $L_p$ baselines. Selected $L_p$ models are the best Resnet50s from the bench-marking done in \cref{fig:UA2-imagenet-medium}, and for computational budget reasons they are trained on a 100-image subset of ImageNet, constructured by taking every 10th class.}
\end{table}

\begin{table}[]

\label{tab:pixmix-adv}
\centering
\begin{tabular}{lrr}
\toprule
 Model & Clean Acc. & UA2 \\
\midrule
WRN-40-2 + PixMix & \textbf{95.1} & 15.00 \\ \midrule
WRN-28-10 + $L_\infty$ 4/255 & 89.3 & 37.3 \\
WRN-28-10 + $L_\infty$ 4/255 + PixMix & 91.4 & \textbf{45.1} \\ \midrule
WRN-28-10 + $L_\infty$ 8/255 & 84.3 & 41.4 \\
WRN-28-10 + $L_\infty$ 8/255 + PixMix & 87.1 & \textbf{47.4}\\
\bottomrule \\
\end{tabular}
\caption{\textbf{PixMix and $L_p$ training.} We compare UA2 performance on CIFAR-10 of models trained with PixMix and adversarial training. Combining PixMix with adversarial training results in large improvements in $\UAA$, demonstrating an exciting future direction for improving unforeseen robustness. All numbers denote percentages, and $L_\infty$ training was performed with the TRADES algorithm.}
\end{table}

\section{Human Study of Semantic Preservation}
\label{human-study}

\begin{table}[!ht]
\label{tab:human-studies-table}
\centering
\begin{center}
\begin{small}
\begin{tabular}{lcc}\toprule
Attack Name & Correct & Corrupted or Ambiguous \\ \midrule
Clean        &95.4 &4.2 \\\midrule
Elastic &92.0 &2.0 \\
Gabor &93.4 &4.0 \\
Glitch &80.2 &16.0 \\
JPEG &93.4 & 0.6 \\
Kaleidescope &93.0 &6.2 \\
Pixel &92.6 &1.8 \\
Snow &90.0 &3.2 \\
Wood &91.4 &1.8 \\ \midrule
Adversarial images average & 91.2 & 4.5 \\
\bottomrule\\
\end{tabular}
\caption{\textbf{Results of user study.} We run a user study on the 200 class subset of ImageNet presented as part of ImageNet-R \citep{hendrycks2021many},  assessing the multiple-choice classification accuracy of human raters, allowing raters to choose certain images as corrupted. We use 4 raters per label and take a majority vote, finding high classification accuracy across all attacks.}
\end{small}
\end{center}
\end{table}

We ran user studies to compare the difficulties of labeling the adversarial examples compared to the clean examples.  We observe that under our distribution of adversaries users experience a 4.2\% drop in the ability to classify. This highlights how overall humans are still able to classify over 90\% of the images, implying that the attacks have not lost the semantic information, and hence that models still have room to grow before they match human-level performance on our benchmark.

In line with ethical review considerations, we include the following information about our human study:

\begin{itemize}
    \item \textbf{How were participants recruited?} We made use of the \href{https://www.surge.ai/}{\textcolor{blue}{\underline{surgehq.ai}}} platform to recruit all participants.
    \item \textbf{How were the participants compensated?} Participants were paid at a rate of $\$0.05$ per label, with an average rating time of 4 seconds per image---ending at an average rate of roughly $\$45$ hour.
    \item \textbf{Were participants given the ability to opt out?} All submissions were voluntary.
     \item \textbf{Were participants told of the purpose of their work?} Participants were told that their work was being used to "validate machine learning model performance".
    \item \textbf{Was any data or personal information collected from the participants?} No personal data was collected from the participants.
    \item \textbf{Was there any potential risks done to the participants?} Although some ImageNet classes are sometimes known to contain elicit or unwelcome content \citet{Prabhu2019TheBD}. Our 100-class subset of ImageNet purposefully excludes such classes, and as such participants were not subject to any undue risks or personal harms.
    
\end{itemize}

\newpage

\begin{figure}[htbp]
\centering
\includegraphics[width=\textwidth]{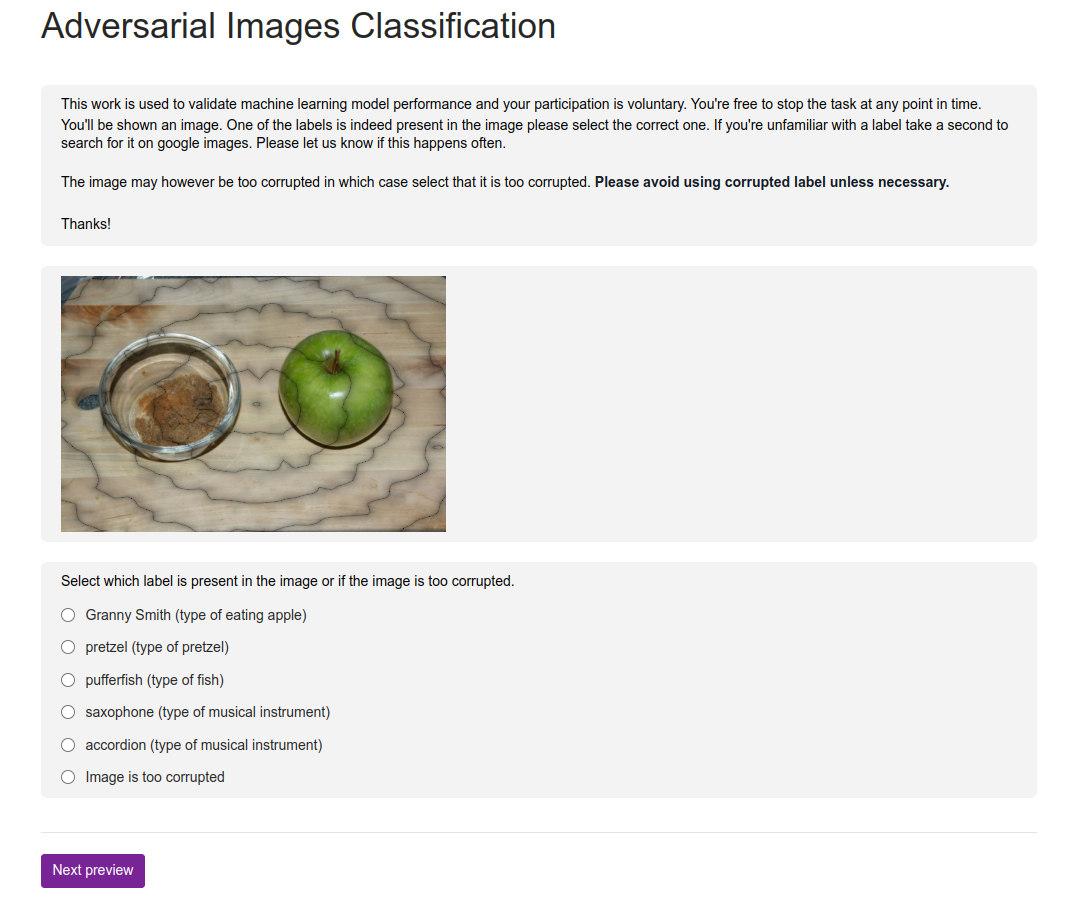}
\caption{\textbf{Interface of participants.} We demonstrate the interface which was provided to the participants of the study, involving the selection of correct classes from our 100-class subset of ImageNet. }
\label{fig:participant_interface}  
\end{figure}

\begin{figure}[h]
\begin{minipage}{\textwidth}
\footnotesize
{ 
\fontfamily{qcr}\selectfont
This work is used to validate machine learning model performance and your participation is voluntary. You're free to stop the task at any point in time.

You'll be shown an image. One of the labels is indeed present in the image please select the correct one. If you're unfamiliar with a label take a second to search for it on google images. Please let us know if this happens often.

The image may however be too corrupted in which case select that it is too corrupted. Please avoid using corrupted label unless necessary.
Thanks!
}
\end{minipage}
\caption{\textbf{Instructions given to the participants.} Above is a list of the instructions which were given to the participants in the human study.}
\label{fig:your_label}
\end{figure}

\newpage
\section{Correlation of $L_p$ robustness and $\task$}
\label{app:lp-correlation}
\begin{figure}[h]
    \centering
    \includegraphics[width=\linewidth]{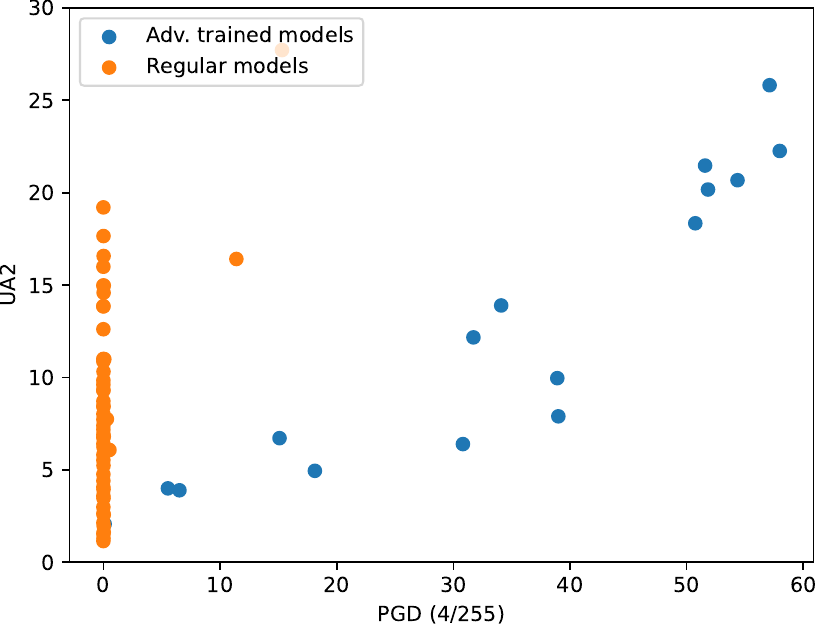}
    \caption{\textbf{$L_p$ robustness correlates with UA2.} Across our benchmark, for adversarially trained models $L_p$ robustness correlates with $\UAA$ - however, several models trained without adversarial training still improve on $\UAA$.}
    \label{fig:lp-correlated-graph}
\end{figure}

\section{Grid search vs. gradient-based search}
\label{app:grid-vs-gradient}
\begin{table}[h!]

\label{tab:gridsearch}
\centering
\begin{tabular}{cc}
\toprule
Optimisation Technique       & UA2   \\  \midrule
Randomized grid search           &  74.1    \\
Gradient-based search (ours)    & 7.2  \\ \bottomrule \\
\end{tabular}
\caption{\textbf{Comparing gradient-based search to grid-based search} We compare the performance of optimising with a randomised grid-based search using 1000 forward-passes per datapoint, finding that our gradient-based methods perform a lot  better than this compute-intensive baseline. }
\end{table}

\newpage
\section{Transfer Attacks}
\label{app:black-box-transfer}
    \cref{tab:transfer} shows the transfer-attack performances across various source and target models based on 1000 test samples. We observe that while the transfer attacks are not as effective as white-box attacks, they consistently outperform baseline unoptimized attacks where the perturbations are randomly initialized (\cref{tab:nonopt_baseline}).

  \begin{table}[h!]

  \label{tab:transfer}
    \resizebox{\textwidth}{!}{%
    \setlength\tabcolsep{4pt}
    \begin{tabular}{@{}llllllllllll@{}}
    \toprule
                                            & Clean Acc. & PGD& UA2   & JPEG & Elastic & Wood & Glitch & Kal. & Pixel & Snow & Gabor \\ \midrule
        ResNet50 (\textbf{source model})             & 75.2      & 0           & 13.2  & 0    & 22.2    & 30.8 & 10     & 4.3          & 4.8   & 3.1  & 30.4  \\
        ViT-small Patch16 ImageNet1K        & 78.5      & 73.1        & 59.99 & 75   & 62.7    & 69.9 & 46     & 48           & 62.8  & 55.5 & 60    \\
        ConvNeXt-V2-tiny ImageNet1K         & 82.1      & 74.8        & 67.66 & 77.1 & 69      & 75.9 & 54     & 60           & 73.6  & 65.2 & 66.5  \\
        Swin-small ImageNet1K +$ L_\infty$ 4/255 & 71.1      & 70.6        & 50.39 & 70.9 & 56.7    & 65.8 & 34.8   & 10.7         & 59.3  & 48.4 & 56.5  \\ \midrule

        ResNet50                                    & 75.2      & 67.9        & 43.19 & 70.1 & 53.1    & 57.7 & 30.1   & 5.4          & 53.3  & 38.1 & 37.7  \\
        ViT-small Patch16 ImageNet1K (\textbf{source model}) & 78.5      & 0           & 6.51  & 0    & 8.2     & 12.7 & 0.5    & 4.7          & 2.1   & 0.8  & 23.1  \\
        ConvNeXt-V2-tiny ImageNet1K                 & 82.1      & 75.7        & 67.3  & 78.6 & 68.5    & 72.8 & 56.4   & 59.9         & 70.1  & 65.1 & 67    \\
        Swin-small ImageNet1K + $ L_\infty$ 4/255              & 71.1      & 70.5        & 50.11 & 70.9 & 57.1    & 65.1 & 35     & 10.8         & 59.5  & 48   & 54.5  \\ \midrule
        
        ResNet50                                   & 75.2      & 67.8        & 42.06 & 68.3 & 51      & 55.7 & 31.7   & 5.8          & 51.7  & 32.1 & 40.2  \\
        ViT-small Patch16 ImageNet1K               & 78.5      & 74.7        & 57.31 & 75   & 60      & 69   & 42     & 46.8         & 57.2  & 50.2 & 58.3  \\
        ConvNeXt-V2-tiny ImageNet1K (\textbf{source model}) & 82.1      & 0           & 12.15 & 0    & 23.2    & 22.3 & 7.4    & 3.5          & 6     & 0.6  & 34.2  \\
        Swin-small ImageNet1K + $ L_\infty$ 4/255             & 71.1      & 71.2        & 50.1  & 71.2 & 56.1    & 65   & 37.8   & 10.7         & 59.1  & 45   & 55.9  \\ \midrule
        
        ResNet50                                      & 75.2      & 64          & 36.95 & 61.8 & 42.5    & 57.8 & 15.6   & 5.4          & 45.3  & 29.2 & 38    \\
        ViT-small Patch16 ImageNet1K                  & 78.5      & 66.9        & 53.3  & 70.6 & 51.4    & 68.2 & 23.8   & 47.1         & 58.4  & 44.2 & 62.7  \\
        ConvNeXt-V2-tiny ImageNet1K                   & 82.1      & 75.5        & 65.26 & 75.7 & 64.7    & 74.5 & 46.1   & 58.2         & 72.3  & 63.6 & 67    \\
        Swin-small ImageNet1K + $ L_\infty$  4/255 (\textbf{source model}) & 71.1      & 53.8        & 21.4  & 42   & 17.9    & 42.3 & 5.1    & 5.1          & 7.6   & 3.4  & 47.8  \\ \bottomrule        
    \end{tabular}%
    }
      \caption{Transfer attack performance}
  \end{table}

\begin{table}[h]

\label{tab:nonopt_baseline}
\resizebox{\textwidth}{!}{%
\setlength\tabcolsep{4pt}
\begin{tabular}{@{}llllllllllll@{}}
\toprule
                               & Clean Acc. & PGD & UA2   & JPEG & Elastic & Wood & Glitch & Kal. & Pixel & Snow & Gabor \\ \midrule
ResNet50                       & 75.2      & 74.1        & 56.44 & 74.3 & 62.8    & 55.7 & 55.8   & 6.3          & 74.1  & 74.8 & 47.7  \\
ViT-small Patch16 ImageNet1K   & 78.5      & 78          & 69.19 & 78   & 70.2    & 70.2 & 65.4   & 47.7         & 77.3  & 78.6 & 66.1  \\
ConvNeXt-V2-tiny ImageNet1K    & 82.1      & 82.2        & 74.74 & 82.2 & 75.2    & 74.4 & 69.7   & 60.7         & 81.5  & 81.4 & 72.8  \\
Swin-small ImageNet1K + $ L_\infty$  4/255 & 71.1      & 71.3        & 58.19 & 71.6 & 62      & 63.4 & 58     & 10.2         & 70.9  & 71.7 & 57.7  \\ \bottomrule
\end{tabular}%
}
\caption{Unoptimized attack performance}
\end{table}

\section{X-Risk Sheet}
We provide an analysis of how our paper contributes to reducing existential risk from AI, following the framework suggested by \citet{hendrycks2022x}. Individual question responses do not decisively imply relevance or irrelevance to existential risk reduction.

\subsection{Long-Term Impact on Advanced AI Systems}
In this section, please analyze how this work shapes the process that will lead to advanced AI systems and how it steers the process in a safer direction.

\begin{enumerate}
\item \textbf{Overview.} How is this work intended to reduce existential risks from advanced AI systems? \\
\textbf{Answer:} This work explores robustness of neural networks to unforeseen forms of optimization pressure. Advanced AI systems may be highly effective and creative optimizers, capable of carrying out ``zero-day'' attacks on software systems and other AIs alike. Improving the robustness of AIs to unforeseen attacks may protect them against powerful adversaries seeking to break them. In some cases, this could reduce existential risk. For example, biothreat screening tools could leverage classifiers that are robust to unforeseen attacks to resist highly advanced attempts at evading detection. Additionally, neural network proxy objectives that lack robustness to optimization pressure could lead to catastrophic outcomes if optimized to an extreme degree \citep{hendrycks2023overview}.

\item \textbf{Direct Effects.} If this work directly reduces existential risks, what are the main hazards, vulnerabilities, or failure modes that it directly affects? \\
\textbf{Answer:} This work directly reduces risks from proxy gaming and bioterrorism (via improved robustness of screening tools).

\item \textbf{Diffuse Effects.} If this work reduces existential risks indirectly or diffusely, what are the main contributing factors that it affects? \\
\textbf{Answer:} By focusing on unforeseen attacks, or work encourages a security mindset that recognizes a multitude of potential vulnerabilities, including ones that have not been considered yet. By proposing a safety benchmark, we hope to improve safety culture and the amount of safety research in the ML community.

\item \textbf{What’s at Stake?} What is a future scenario in which this research direction could prevent the sudden, large-scale loss of life? If not applicable, what is a future scenario in which this research direction be highly beneficial? \\
\textbf{Answer:} Malicious actors could use advanced AIs to help them develop bioweapons that evade most screening and detection mechanisms. Adversarially robust detectors are crucial for mitigating this risk, and robustness to unforeseen attacks is necessary when dealing with advanced AI-assisted design processes.

\item \textbf{Result Fragility.} Do the findings rest on strong theoretical assumptions; are they not demonstrated using leading-edge tasks or models; or are the findings highly sensitive to hyperparameters? \hfill
$\square$
\item \textbf{Problem Difficulty.} Is it implausible that any practical system could ever markedly outperform humans at this task? \hfill $\square$
\item \textbf{Human Unreliability.} Does this approach strongly depend on handcrafted features, expert supervision, or human reliability? \hfill $\square$
\item \textbf{Competitive Pressures.} Does work towards this approach strongly trade off against raw intelligence, other general capabilities, or economic utility? \hfill $\boxtimes$
\end{enumerate}

\subsection{Safety-Capabilities Balance}
In this section, please analyze how this work relates to general capabilities and how it affects the balance between safety and hazards from general capabilities.

\begin{enumerate}[resume]
\item \textbf{Overview.} How does this improve safety more than it improves general capabilities? 

\textbf{Answer:} We propose a benchmark that enables quantifying differential progress on robustness to unforeseen adversaries relative to clean accuracy. We find that methods improving clean accuracy also improve unforeseen robustness, but some methods do provide differential improvements to UA2, including adversarial training and data augmentation. Adversarial robustness is widely considered to be in tension with clean accuracy. In particular, improving robustness through adversarial training reduces clean accuracy . Thus, developing methods to improve unforeseen robustness is unlikely to substantially improve general capabilities as well.

\item \textbf{Red Teaming.} What is a way in which this hastens general capabilities or the onset of x-risks? 
\textbf{Answer:} Improving the robustness of proxy objectives to optimization pressure could improve the effectiveness of reward-based fine-tuning of AI systems , which would improve general capabilities as well as safety.

\item \textbf{General Tasks.} Does this work advance progress on tasks that have been previously considered the subject of usual capabilities research? \hfill $\square$

\item \textbf{General Goals.} Does this improve or facilitate research towards general prediction, classification, state estimation, efficiency, scalability, generation, data compression, executing clear instructions, helpfulness, informativeness, reasoning, planning, researching, optimization, (self-)supervised learning, sequential decision making, recursive self-improvement, open-ended goals, models accessing the Internet, or similar capabilities? \hfill $\square$

\item \textbf{Correlation with General Aptitude.} Is the analyzed capability known to be highly predicted by general cognitive ability or educational attainment? \hfill $\square$

\item \textbf{Safety via Capabilities.} Does this advance safety along with, or as a consequence of, advancing other capabilities or the study of AI? \hfill $\square$
\end{enumerate}

\subsection{Elaborations and Other Considerations}
\begin{enumerate}[resume]
\item \textbf{Other.} What clarifications or uncertainties about this work and x-risk are worth mentioning? 

\textbf{Answer:} Regarding Q8, adversarial training reduces clean accuracy while improving robustness to unforeseen adversaries. However, we also find that other methods can improve robustness to unforeseen adversaries without trading off clean accuracy. Thus, there may be ways of achieving high robustness to unforeseen adversaries without trading off significant amounts of clean accuracy. However, these methods may still incur an overhead cost in terms of compute resources.
\end{enumerate}